%% file: main.tex
\documentclass[aps,prl,twocolumn,superscriptaddress]{revtex4}

\usepackage{mathptmx}      
\usepackage{helvet}        
\usepackage{courier}       

\makeatletter
\def\@fnsymbol#1{\ensuremath{\ifcase#1\or \dagger\or *\or \ddagger\or
   \mathsection\or \mathparagraph\or \|\or **\or \dagger\dagger
   \or \ddagger\ddagger \else\@ctrerr\fi}}
\makeatother

\usepackage{dcolumn}
\usepackage{amsfonts}
\usepackage{amsmath}
\usepackage{amssymb}
\usepackage{amsthm}
\usepackage{mathrsfs}
\usepackage{float}
\usepackage{amsbsy}
\usepackage{amstext}
\usepackage{siunitx}
\usepackage{fixltx2e}
\usepackage{upgreek,xspace}
\usepackage{multirow}
\usepackage{wasysym}
\usepackage{mathtools}
\usepackage{bbold,bm}
\usepackage{graphicx}
\usepackage{booktabs}
\usepackage[colorlinks=true,citecolor=blue,linkcolor=blue,urlcolor=blue]{hyperref}
\usepackage{bm}
\usepackage{caption}
\usepackage{xcolor}
\usepackage{tikz}      
\usepackage{makecell} 
\usepackage{titlesec}
\usepackage{ragged2e} 
\usepackage{algorithm}
\usepackage{algorithmic}

\usepackage{pgffor} 
\graphicspath{{supplementary/Figures/scatter_results}} 



\newcommand{\panel}[3][0.48\linewidth]{%
  \begin{tikzpicture}
    \node[inner sep=0] (img) {\includegraphics[width=#1]{#2}};
    \node[anchor=north west, xshift=-15pt, yshift=-6pt, fill=white, rounded corners=2pt, inner sep=2pt, text=black, font=\bfseries] 
         at (img.north west) {#3};
  \end{tikzpicture}%
}

\definecolor{burgundy}{RGB}{128,0,32}  
\definecolor{myDarkGreen}{RGB}{11,61,46}

\titleformat{\section}[block]
  {\normalfont\bfseries\Large}  
  {\thesection.}{0.75em}{}      

\titleformat{\subsection}[block]
  {\normalfont\bfseries\large}
  {\thesubsection.}{0.75em}{}

\titleformat{\subsubsection}[block]
  {\normalfont\bfseries\normalsize}
  {\thesubsubsection.}{0.75em}{}

\titlespacing*{\section}{0pt}{*2}{*1}
\titlespacing*{\subsection}{0pt}{*1.5}{*0.75}
\titlespacing*{\subsubsection}{0pt}{*1.2}{*0.6}

\DeclareCaptionJustification{justified}{\justifying}

\begin{document}

\title{\LARGE Universal Battery Degradation Forecasting Driven by Foundation Model Across Diverse Chemistries and Conditions}

\input{manuscript/information.tex}

\input{manuscript/abstract.tex}

\maketitle

\newpage
\input{manuscript/introduction.tex}

\input{manuscript/results.tex}
\input{manuscript/discussion.tex}
\input{manuscript/method.tex}

\input{manuscript/addition.tex}

\clearpage
\bibliography{mybib.bib}



\end{document}

%% file: manuscript/information.tex
\author{Joey Chan}
\altaffiliation{These authors contributed equally to this work.}
\affiliation{Department of Industrial Engineering and Management, Shanghai Jiaotong University, Shanghai, China.}

\author{Huan Wang}
\altaffiliation{These authors contributed equally to this work.}
\affiliation{Department of Systems Engineering, The City University of Hong Kong, Hong Kong.}

\author{Haoyu Pan}
\affiliation{Department of Industrial Engineering and Management, Shanghai Jiaotong University, Shanghai, China.}

\author{Wei Wu}
\affiliation{Department of Industrial Engineering and Management, Shanghai Jiaotong University, Shanghai, China.}

\author{Zirong Wang}
\affiliation{Department of Industrial Engineering and Management, Shanghai Jiaotong University, Shanghai, China.}

\author{Zhen Chen}
\affiliation{Department of Industrial Engineering and Management, Shanghai Jiaotong University, Shanghai, China.}

\author{Ershun Pan}
\affiliation{Department of Industrial Engineering and Management, Shanghai Jiaotong University, Shanghai, China.}

\author{Min Xie}
\affiliation{Department of Systems Engineering, The City University of Hong Kong, Hong Kong.}

\author{Lifeng Xi}
\affiliation{Department of Industrial Engineering and Management, Shanghai Jiaotong University, Shanghai, China.}

%% file: manuscript/abstract.tex
\begin{abstract}
\noindent \normalsize 
Accurate forecasting of battery capacity fade is essential for the safety, reliability, and long-term efficiency of energy storage systems. 
However, the strong heterogeneity across cell chemistries, form factors, and operating conditions makes it difficult to build a single model that generalizes beyond its training domain. 
This work proposes a unified capacity forecasting framework that maintains robust performance across diverse chemistries and usage scenarios. 
We curate 20 public aging datasets into a large-scale corpus covering 1,704 cells and 3,961,195 charge–discharge cycle segments, spanning temperatures from $-5,^{\circ}\mathrm{C}$ to $45,^{\circ}\mathrm{C}$, multiple C-rates, and application-oriented profiles such as fast charging and partial cycling. 
On this corpus, we adopt a Time-Series Foundation Model (TSFM) backbone and apply parameter-efficient Low-Rank Adaptation (LoRA) together with physics-guided contrastive representation learning to capture shared degradation patterns. 
Experiments on both “seen” and deliberately held-out “unseen” datasets show that a single unified model achieves competitive or superior accuracy compared with strong per-dataset baselines, while retaining stable performance on chemistries, capacity scales, and operating conditions excluded from training. 
These results demonstrate the potential of TSFM-based architectures as a scalable and transferable solution for capacity degradation forecasting in real battery management systems.

\end{abstract}

%% file: manuscript/introduction.tex
\begin{figure*}[htb]
  \centering
  \panel[0.95\linewidth]{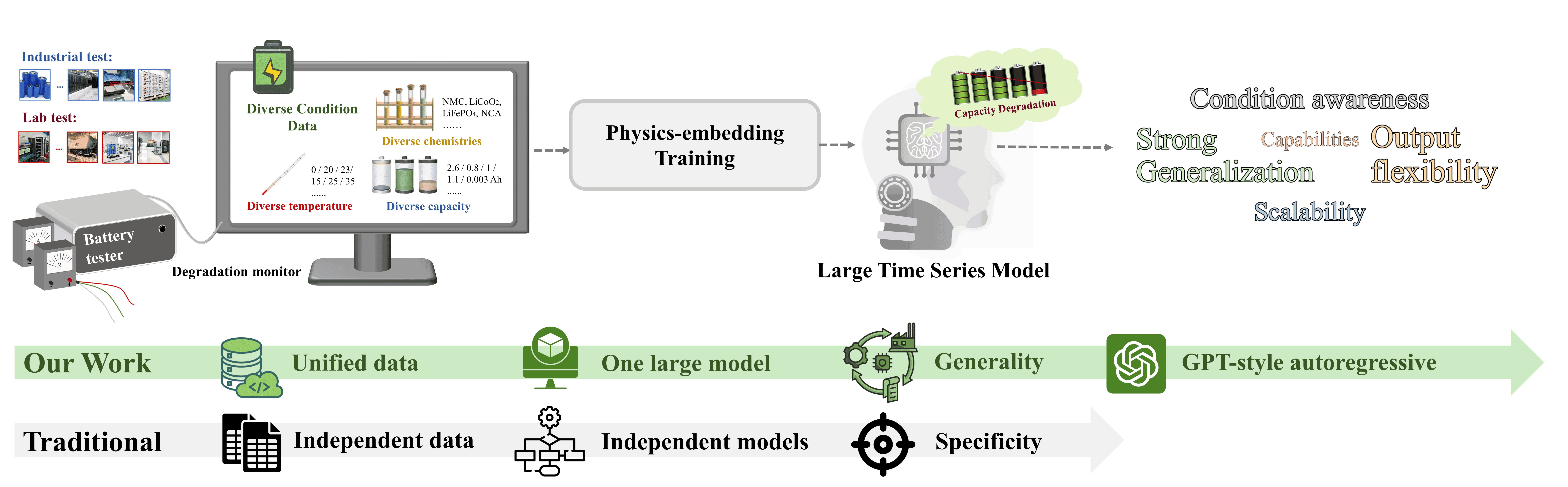}{(a)}\hfill
  \panel[0.95\linewidth]{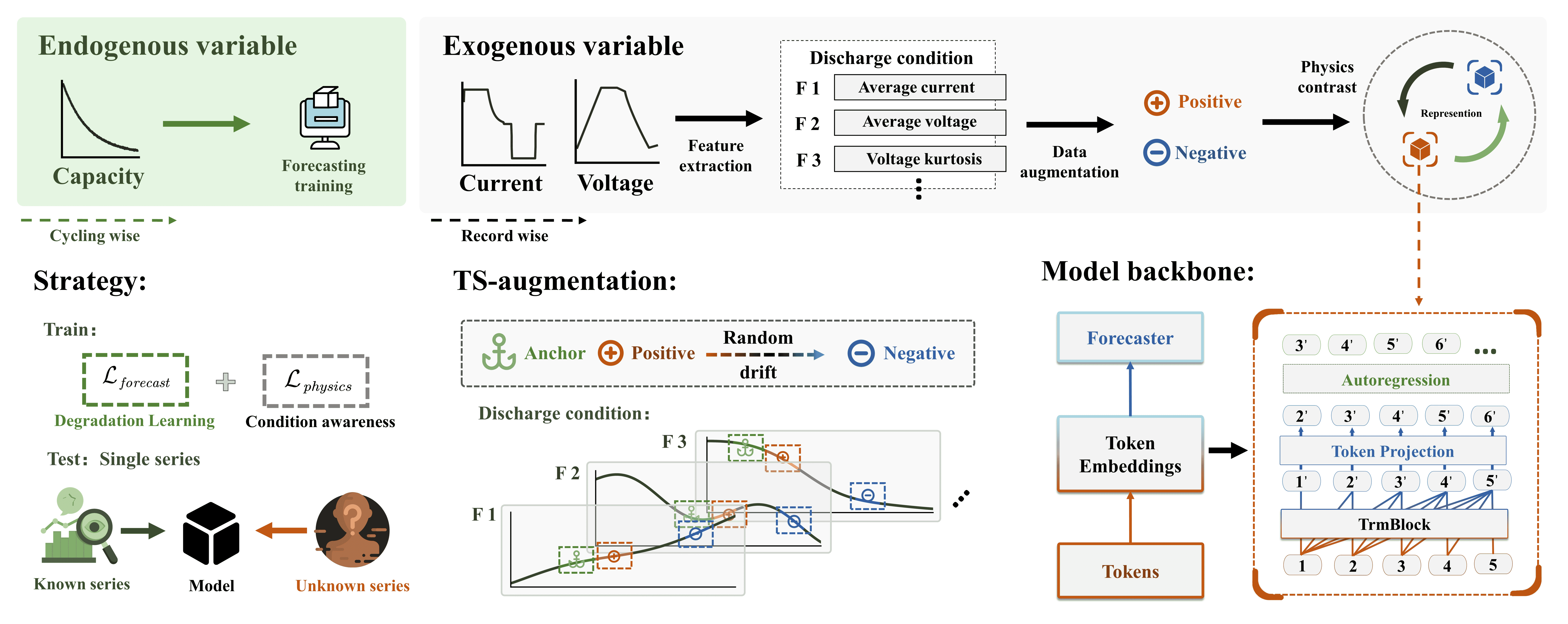}{(b)}
  \caption{\textbf{The flowchart of the large-model-driven battery capacity forecasting.} 
  \textbf{(a)} Unlike conventional, task-specific capacity degradation models, our approach constructs a rich, standardized corpus of degradation trajectories and employs a single time-series foundation model to forecast capacity under previously unseen conditions. 
  \textbf{(b)} In the physics-informed training strategy, the endogenous variable (capacity) and exogenous variables (operating conditions) are treated differently: 
  charge–discharge features are incorporated as auxiliary physical embeddings during training to exploit domain knowledge, while the model remains independent of these potentially missing physical inputs at inference time.}
  \label{overview}
\end{figure*}

Lithium-ion batteries have become a central pillar of modern energy infrastructures, supporting applications that range from electric vehicles (EVs) and portable electronics to grid-scale storage, frequency regulation, and distributed renewable integration. In 2023, global EV sales exceeded 14 million units, corresponding to a battery demand of more than 1~TWh and a projected global demand of 3–6~TWh by 2030 \cite{radha2025sustainable,link2025feasibility}. 
The global battery market was valued at approximately USD~134.6~billion in 2024 and is expected to reach USD~329.8~billion by 2030, with a compound annual growth rate of about 16.4\% \cite{grandview2024market}. 
In parallel with this rapid expansion, the landscape of battery chemistries is becoming increasingly diverse, including lithium iron phosphate (LFP), nickel cobalt manganese/aluminum (NCM/NCA), sodium-ion, solid-state chemistries, and hybrid systems, deployed across cylindrical, pouch, prismatic, and coin formats with widely varying energy densities, thermal stability, and charging characteristics \cite{cong2024review}. 
For instance, the market share of LFP has risen sharply from about 6\% in 2020 to roughly 30\% in 2022, while NCM/NCA systems continue to dominate long-range EV applications due to their high specific energy \cite{dini2024review}. 
This technological diversification places stringent new demands on battery management systems (BMSs), especially on capacity degradation models that must operate reliably across chemistry families, form factors, and operating regimes \cite{zou2024leaching,liu2019challenges}.

Accurate forecasting of capacity degradation is a key enabler for safe, reliable, and cost-effective operation over the entire lifetime of energy storage assets. 
However, capacity fade is simultaneously governed by chemistry-specific reactions, electrode–electrolyte interfaces, cell design, and application-dependent load patterns, which together give rise to highly heterogeneous temporal behaviors across use cases. 
Batteries deployed in EVs, grid-scale BESS, and consumer electronics experience fundamentally different cycling depths, current profiles, temperature environments, and calendar aging effects, often extending far beyond the controlled conditions of laboratory tests \cite{lu2023deep,zhang2023realistic}. 
The gap between laboratory protocols and real-world duty cycles means that models tuned for specific datasets, chemistries, or temperature ranges can suffer severe performance degradation when transferred to unseen devices or operating conditions. 
This problem is further exacerbated by the emergence of new chemistries such as sodium-ion \cite{tan2025batterylife} and mixed-chemistry systems \cite{zhu2022data}, which introduce additional and sometimes interacting degradation mechanisms. 
As a result, the central challenge is no longer merely to improve accuracy in a single, well-controlled scenario, but to build a capacity forecasting framework that can generalize across chemistries, structures, and usage conditions—and remain robust when encountering previously unseen regimes.

Hybrid modeling approaches that combine physics-informed indicators with data-driven techniques have shown promise in improving interpretability and prediction accuracy \cite{zhang2020identifying}. 
Nevertheless, their generalization often remains tightly coupled to hand-crafted features and task-specific architectures. 
When the target deviates from the training domain—for example, when the chemistry changes from LCO to NMC, the temperature range shifts, or the duty cycle becomes more dynamic—these models still tend to exhibit noticeable performance drops \cite{zhang2023realistic}. 
From a broader perspective, capacity forecasting is fundamentally a time-series prediction problem: given a historical trajectory of capacity (and possibly auxiliary signals), the task is to infer its future evolution. 
This formulation aligns closely with other sequence modeling problems such as language modeling, where future tokens are generated conditionally on past context. 
This conceptual similarity suggests that the same architectural tools that have transformed natural language processing—most notably Transformer-based models—may provide a powerful and more universal modeling paradigm for battery degradation.

Transformer architectures \cite{vaswani2017attention}, originally developed for large language models (LLMs), have recently been adapted to time-series tasks by replacing discrete token embeddings with continuous numerical segments and redefining the training objective from next-token prediction to multi-horizon forecasting \cite{liu2024timer}. 
Time-Series Foundation Models (TSFMs) push this idea further by pretraining large, attention-based models on diverse temporal datasets so that they learn general-purpose temporal priors that can later be adapted to a variety of downstream tasks. 
In principle, such foundation models could provide a unified backbone for battery capacity forecasting: instead of designing bespoke models for each dataset or chemistry, one could leverage a pre-trained TSFM and specialize it to degradation trajectories via lightweight adaptation. 
This raises an important research question: can a TSFM-based framework, suitably adapted to battery domain characteristics, break the long-standing generalization bottlenecks across chemistries, form factors, and operational conditions?

Building on this understanding, this study proposes a unified battery capacity forecasting framework that operates across multiple chemistries and operating conditions. 
We first systematically integrate and standardize 20 publicly available battery aging datasets to construct a large-scale corpus of capacity degradation time series comprising 1,704 cells, covering both lithium-ion and sodium-ion systems, including LCO, NCM, and NCA cathode chemistries as well as cylindrical, pouch, prismatic, and coin cell formats. 
This corpus spans a wide temperature range from $-5\,^{\circ}\mathrm{C}$ to $45\,^{\circ}\mathrm{C}$ and includes complex operating regimes such as fast charging, partial cycling, and dynamic load profiles, thereby providing a solid data foundation for unified modeling.

On this basis, we adopt a time-series foundation model as the forecasting backbone and develop a unified representation of capacity degradation dynamics through parameter-efficient adaptation and physics-guided representation learning. 
Experimental results show that the proposed method achieves stable and accurate capacity forecasts across a variety of challenging scenarios, while maintaining consistent predictive performance for battery chemistries and operating conditions that are unseen during training. 
These findings indicate that our approach provides a scalable and transferable modeling paradigm for battery capacity degradation forecasting, establishes a methodological foundation for deploying time-series foundation models in practical battery management systems, and opens a new research avenue for intelligent battery health prediction under previously unseen operating conditions.

%% file: manuscript/results.tex
\section*{Results}
\subsection*{Framework overview.}
We propose a unified modeling framework for battery capacity degradation forecasting (\textbf{Fig.} \ref{overview}), designed to overcome the performance bottlenecks of traditional data-driven methods under diverse physico-chemical operating conditions.
By incorporating large-scale time-series modeling techniques, the framework enables generalized degradation prediction across various battery chemistries, packaging formats, and application scenarios.

\textbf{Fig.}~\ref{overview}(a) illustrates the motivation and overall workflow of the proposed approach. 
To construct a time-series corpus of battery cycle–capacity degradation trajectories for TSFM training, we aggregate 20 datasets collected from both laboratory experiments and production environments. 
These datasets cover a wide range of chemistries, test temperatures, cell formats, and nominal capacities. 
Using the proposed physics-informed training strategy, this rich degradation corpus exposes the TSFM to sufficiently diverse operating scenarios. 
Our aim is that, given adequate data, Transformer-based capacity forecasting can exhibit a level of universality, generalization, and scalability analogous to that of LLMs in text generation.

\textbf{Fig.}~\ref{overview}(b) provides a more detailed view of the procedure. In our formulation, the sequence of capacity values evolving with cycle index is treated as the endogenous variable. 
In parallel, the training data record the charge–discharge behavior within each cycle; 
from these profiles we extract representative physics-based features, which serve as auxiliary exogenous variables. 
At the model level, we adopt a hybrid objective. On one hand, the capacity sequence is used as the primary target and is fed into the Timer architecture for supervised learning. 
On the other hand, the multidimensional physical features are embedded as auxiliary vectors and optimized via a contrastive learning objective. 
Through data augmentation, this strategy constructs positive and negative sample pairs, encouraging the model to automatically discriminate degradation patterns under different operating conditions and to develop a self-supervised representation of the underlying physical states.
Importantly, the exogenous variables are only used during training; no physical inputs are required at inference, thereby preserving the flexibility and applicability of the model in real-world deployment. Additional methodological details are provided in the Methods section.

\subsection*{Data generation.}

\input{manuscript/Tables/datasets.tex}

\begin{figure*}[!htbp]
  \centering
  \panel[0.95\linewidth]{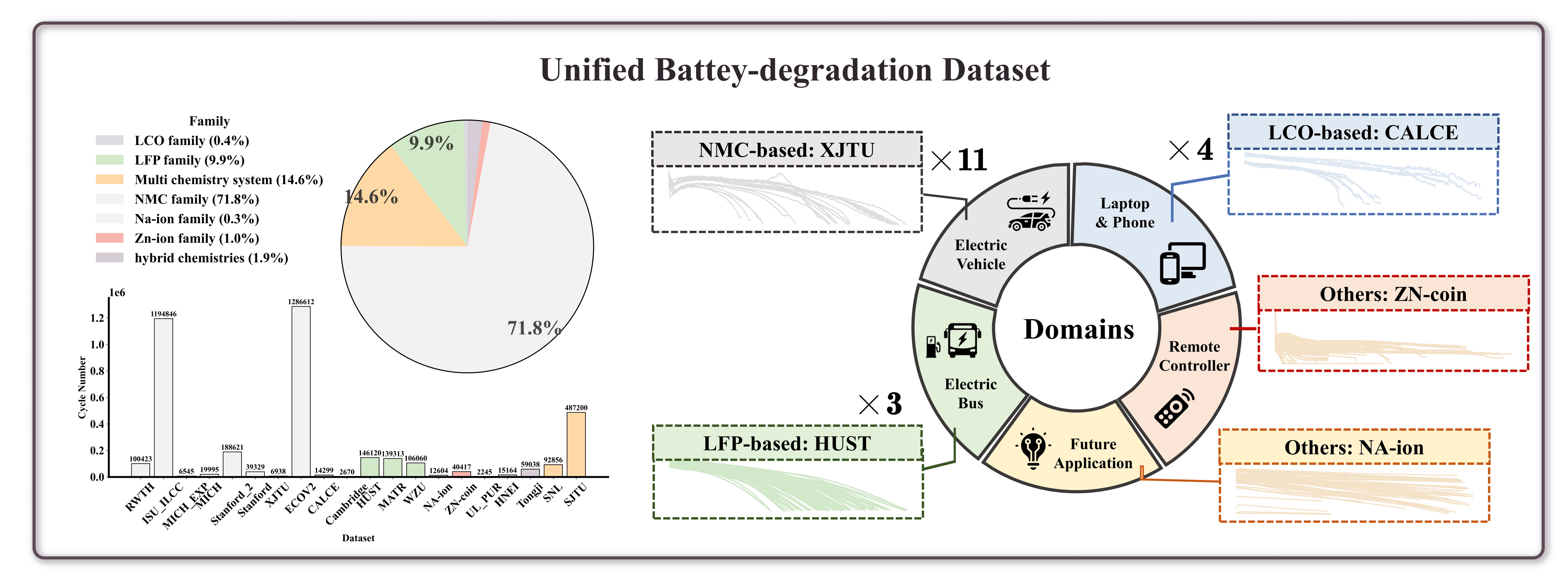}{(a)}\hfill
  \panel[0.95\linewidth]{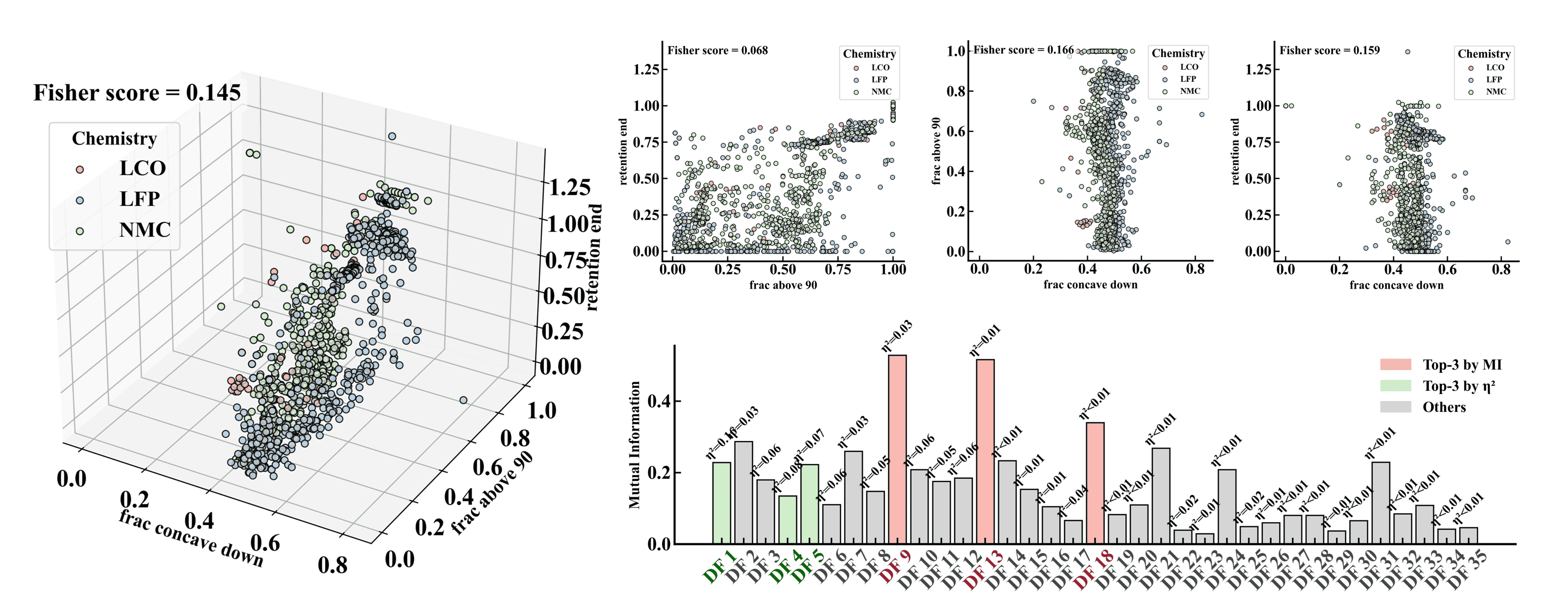}{(b)}
    \panel[0.95\linewidth]{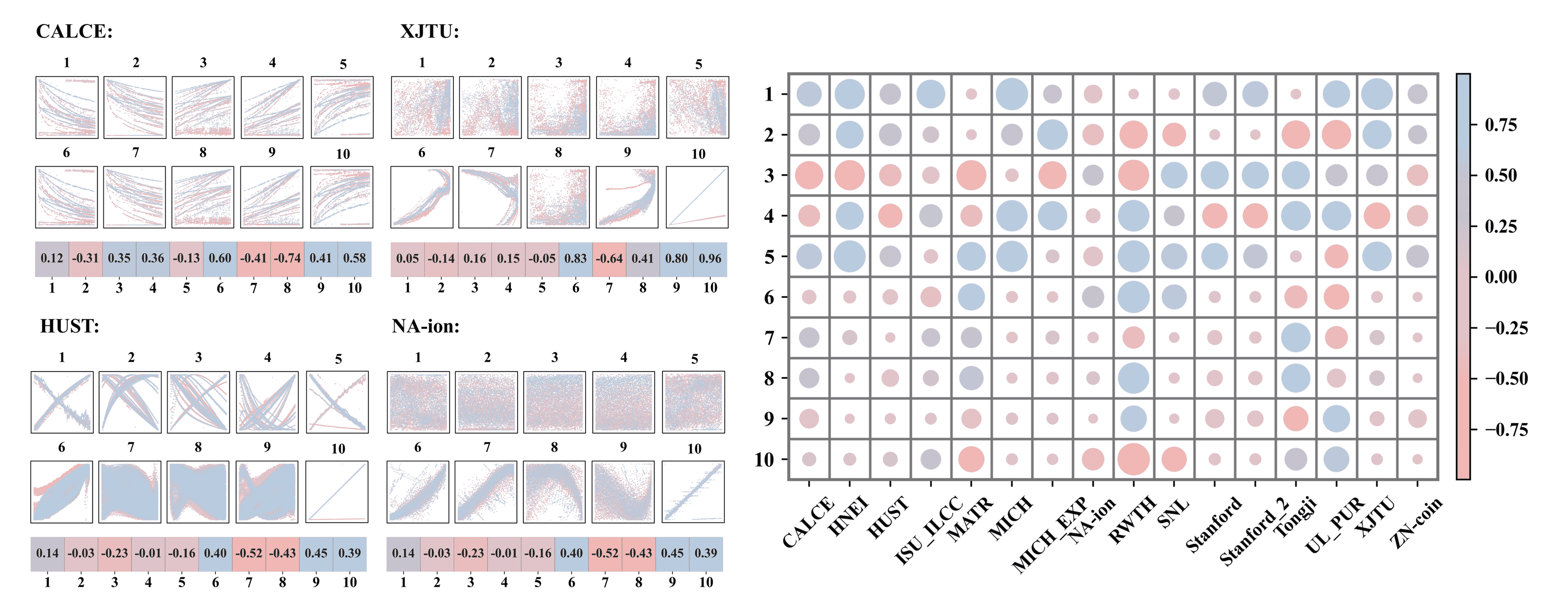}{(c)}
  \caption{\textbf{Dateset overview.} 
  \textbf{(a)}  Overview of the unified battery degradation corpus constructed in this study.
The integrated dataset spans a broad range of electrochemical systems and application scenarios; 
NMC-based cells account for the vast majority of samples, while emerging sodium-ion, zinc-ion, and mixed-chemistry batteries occupy only a small fraction of the corpus. 
  \textbf{(b)} Representative statistical analysis of DFs with respect to battery chemistry. 
  The panel shows 3D scatter plots, 2D projections, and the corresponding MI and ANOVA effect sizes ($\eta^2$) between DFs and chemistry labels.
  \textbf{(c)} Correlation-coefficient heat map and scatter plots between physical features and capacity (x-axis: physical features; y-axis: capacity). 
  Specifically, we compute ten physical descriptors, including charging time, the mean and variance of current and voltage, voltage kurtosis, and related statistics.}
  \label{data}
\end{figure*}

To train the proposed large time-series model for capacity degradation forecasting, we first construct a unified corpus of cycle--capacity trajectories, as illustrated in \textbf{Fig.}~\ref{data}(a). 
Specifically, we consolidate 20 publicly available battery aging datasets into a single mixed-chemistry corpus comprising 1{,}704 cells and a total of 3{,}961{,}195 recorded cycles (\textbf{Supplementary Fig.} S1). 
For each cell, we extract the sequence of cycle indices and the corresponding discharge capacity, yielding standardized cycle--capacity degradation histories that can be directly used for TSFM training.
A detailed summary of the datasets is provided in Table~\ref{tab:dataset_summary_reclassified}.

In terms of form factor, the corpus spans most formats used in both laboratory experiments and industrial deployments. 
It includes widely adopted 18650 cylindrical cells (e.g., RWTH\cite{li2021one} , HUST\cite{ma2022real} , HNEI\cite{devie2018intrinsic}, MATR\cite{severson2019data}), pouch cells (e.g., MICH \cite{weng2021predicting}, ISU\_ILCC \cite{li2024predicting}), prismatic cells for commercial applications (e.g., CALCE\cite{xing2013ensemble}), and coin cells used for fundamental electrochemical studies (e.g., ZN-coin\cite{tan2025batterylife} and Cambridge\cite{zhang2020identifying}). 
This diversity in cell design ensures that the model is exposed to degradation behaviors typical of both small-format cells and larger, application-oriented devices.

From the perspective of electrode chemistry, the corpus covers both mainstream and emerging material systems. 
Cathode chemistries include lithium iron phosphate (LFP), lithium cobalt oxide (LCO), and several nickel-rich oxide families (e.g., NMC and NCA), while anodes range from natural and artificial graphite to graphite--silicon composites and metallic zinc for aqueous systems. 
In addition, some datasets involve cells with electrolyte additives (e.g., VC-enhanced formulations) or sodium-ion batteries with partially undisclosed chemistries, introducing further compositional variability into the training corpus.

The datasets also exhibit substantial diversity in temperature conditions. Overall, the test temperatures range from approximately $-5\,^{\circ}\mathrm{C}$ to $45\,^{\circ}\mathrm{C}$, encompassing low-temperature stress scenarios as well as high-temperature accelerated aging. 
Beyond controlled thermal chambers, datasets such as WZU were collected under seasonal ambient conditions, thereby reflecting realistic environmental fluctuations including diurnal and seasonal temperature swings. 
Several datasets further provide measurements at multiple temperature levels within the same chemistry, enabling the TSFM to learn how thermal conditions modulate degradation dynamics.

In terms of usage scenarios, the corpus covers a broad spectrum from idealized laboratory protocols to application-oriented cycling. 
It includes standard constant-current/constant-voltage (CC--CV) aging tests, as well as fast-charging profiles, multi-stage variable loads, partial cycling, and other complex duty cycles.
Of particular note, the SJTU dataset was acquired by our group on large-format IFR32135 cells designed for stationary energy storage. 
Its high nominal capacity and ESS-oriented cycling protocols, detailed in the \textbf{Supplementary Fig.} S2, \textbf{Table} S1, complement the predominantly small-capacity cells in other datasets and fill an important gap in the overall corpus. 
Collectively, this unified, richly diversified degradation corpus provides a robust foundation for training a generalizable TSFM that can operate across chemistries, form factors, temperatures, and real-world usage conditions.

\subsection*{Feature extration.}

Feature extraction in this study is divided into two components: analysis of the endogenous characteristics of the degradation trajectories and extraction of exogenous variables from the charge–discharge process.

Although time-series forecasting is a mature research field, the single-series setting often causes the intrinsic structure of the endogenous variable itself to be overlooked. 
To date, no study has systematically examined, across sufficiently diverse scenarios, the endogenous characteristics of capacity degradation during cycling for different battery chemistries. 
Here, we seek to characterize how electrode chemistry shapes the intrinsic behavior of capacity fade. 
Specifically, we first extract degradation features (DFs) from the cycle–capacity trajectories of three representative chemistries: LFP, LCO, and NMC (Detailed in \textbf{Supplementary note}~4). 
As shown in \textbf{Fig.}~\ref{data}(b),S3, after standardization the DFs exhibit high mutual information (MI) with chemistry labels but low ANOVA effect size ($\eta^2$), indicating that different chemistries are primarily distinguished by differences in distributional shape rather than by shifts in the mean. 
Representative scatter plots for selected features further show that the degradation trajectories of different chemistries are not completely disjoint. 
Even when cells with different chemistries are cycled under a wide range of test conditions, their degradation features still display common underlying patterns. 
This shared structure in the endogenous variable provides a theoretical basis for developing TSFM-based capacity-forecasting models that generalize across operating conditions.

During training, our goal is to enhance the model's ability to recognize embedded differences in physical operating conditions, thereby improving the generalization performance of capacity forecasts. 
Prior work such as TimeXer~\cite{wang2024timexer} has demonstrated that appropriately incorporating covariate embeddings can substantially improve forecasting accuracy for the target series. 
However, most existing feature-extraction schemes are designed for specific datasets or charge–discharge protocols and thus lack cross-chemistry generality. 
To address this limitation, we leverage the diversity of our corpus and focus on extracting physical features within each discharge cycle. 
As illustrated in \textbf{Fig.}~\ref{data}(c),S4,S5 the scatter relationships between these physical features and capacity exhibit pronounced chemistry-dependent differences (\textbf{Supplementary note}~5). 
At the corpus level, our analysis corroborates the conclusion of Wang \textit{et al.}~\cite{wang2024physics} that charge–discharge physical features are relatively insensitive to cycling protocols yet more strongly governed by the underlying chemistry. 
This complex dependence between endogenous and exogenous variables motivates the incorporation of physics-based features into our time-series capacity-forecasting framework.

\subsection*{Battery Capacity forecasting.}

\begin{figure*}[!htbp]
  \centering
  \panel[0.47\linewidth]{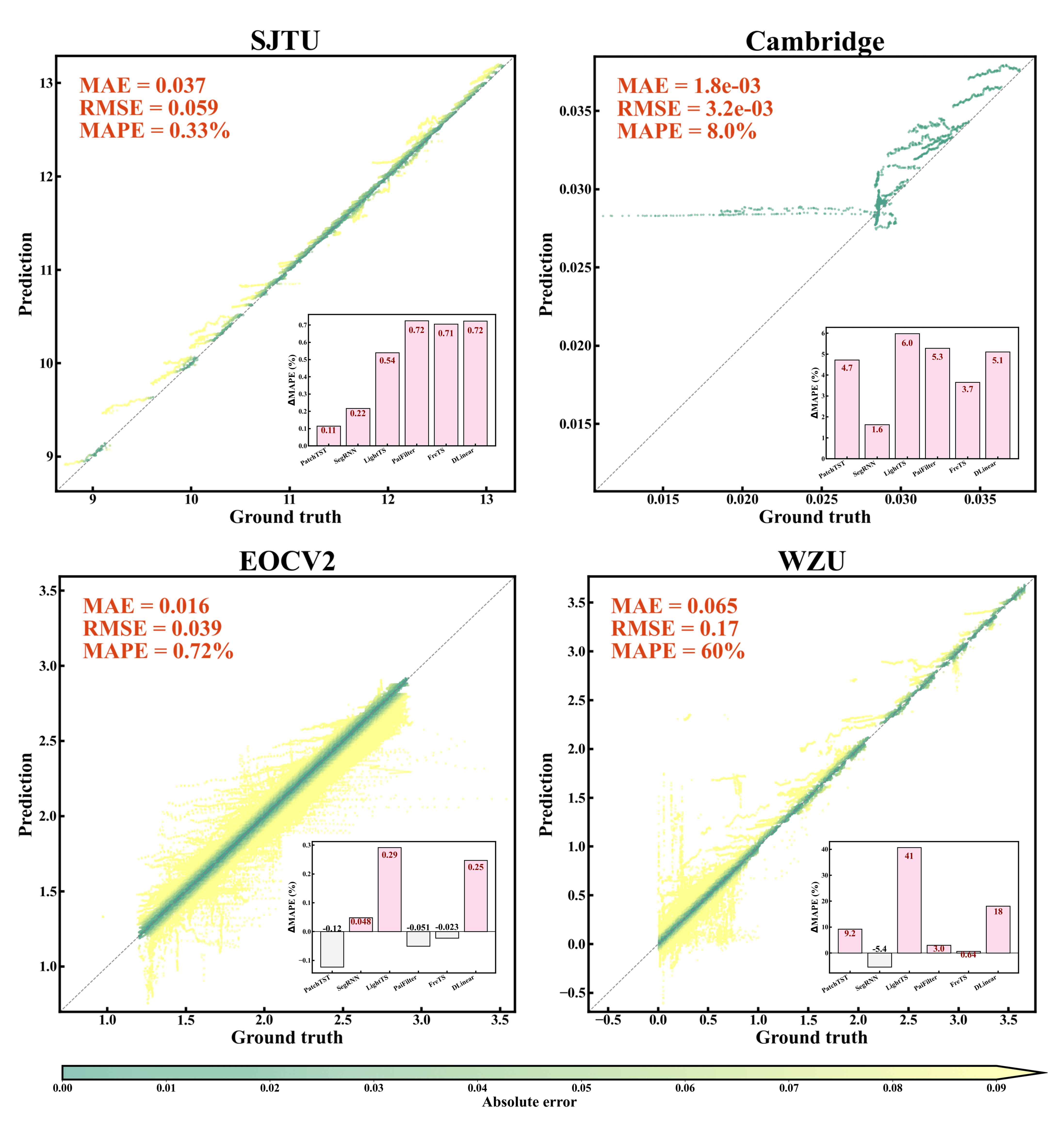}{(a)}\hfill
  \panel[0.47\linewidth]{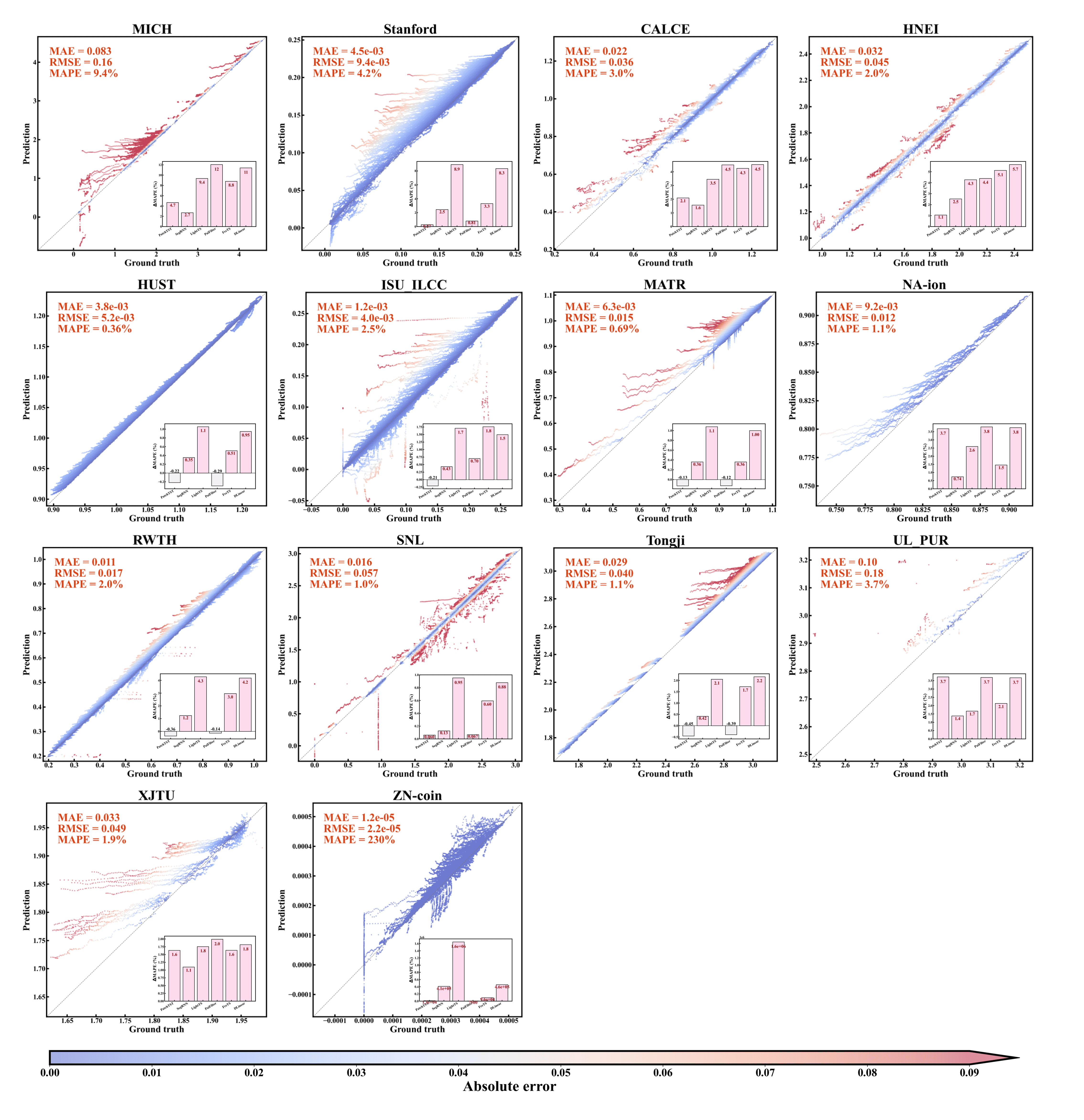}{(b)}\\
    \panel[0.47\linewidth]{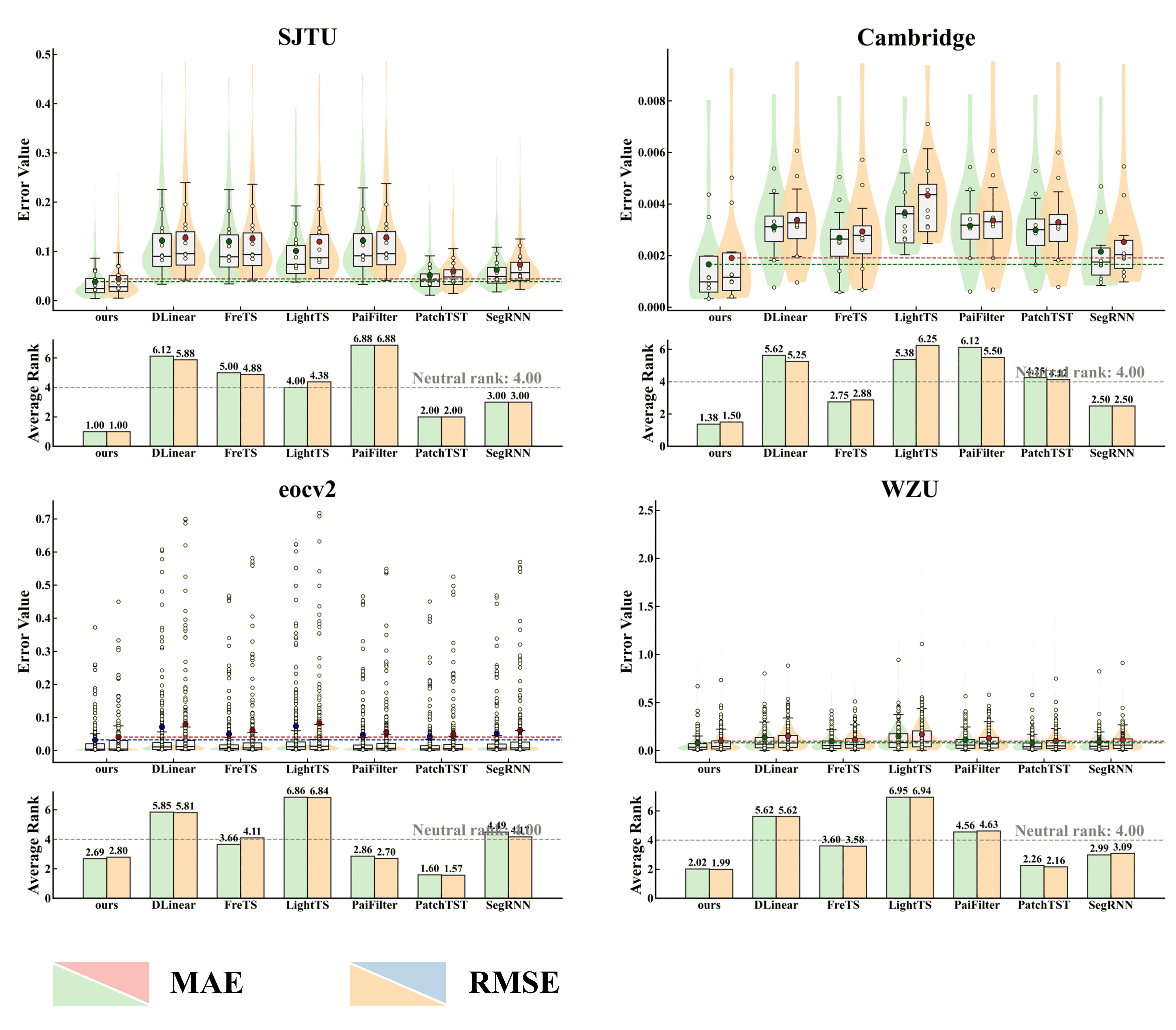}{(c)}\hfill
  \panel[0.47\linewidth]{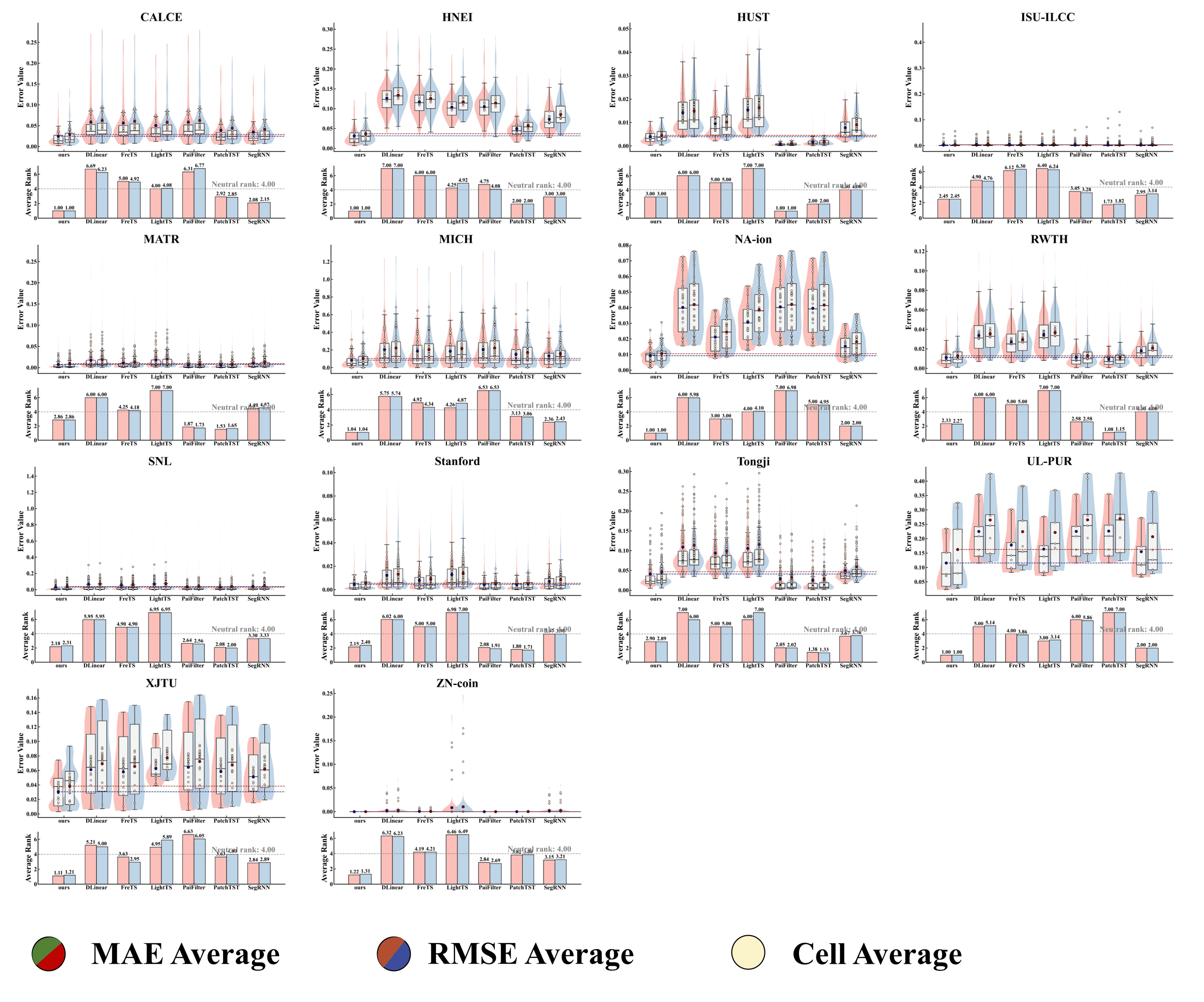}{(d)}
  \caption{\textbf{The illustrations of capacity forecasting results.} 
  \textbf{(a)} Capacity forecasting performance of the proposed method on unseen datasets,
together with the MAPE differences relative to competing baselines, showing that
our approach maintains superior accuracy even under previously unobserved conditions. 
  \textbf{(b)} Capacity forecasting results on seen datasets, where the model exhibits robust
performance across a wide spectrum of battery chemistries, form factors, and
operating scenarios. Using a single time-series foundation model, our method matches
or even surpasses multiple supervised state-of-the-art time-series forecasting
baselines.
\textbf{(c)} Distribution of per-cell average errors and Friedman ranking on unseen datasets.
\textbf{(d)} Distribution of per-cell average errors and Friedman ranking on seen datasets.
The combined box--violin plots visually highlight the advantage of our approach, and
the Friedman rankings show that the proposed single foundation model never falls
outside the top three and attains the best rank on many datasets.
  }
  \label{experiment1}
\end{figure*}

We evaluate the proposed capacity forecasting framework on two regimes: \emph{seen} datasets and \emph{unseen} datasets. 
The seen regime consists of the 16 datasets that participate in model training, whereas the unseen regime contains four datasets---EOCV2, SJTU, WZU, and Cambridge---that are completely excluded during training and only used for testing. 
All methods are compared under a common prediction setting, while the detailed data preprocessing procedures and window-based splitting rules are documented in the in \textbf{Supplementary note}~6. 
It is worth emphasizing that our approach trains \emph{a single} TSFM-based unified model on all seen datasets, whereas each competing method (DLinear\cite{zeng2023transformers}, FreTS\cite{yi2023frequency}, LightTS\cite{zhang2207less}, PaiFilter\cite{yi2024filternet}, PatchTST\cite{nie2022time}, SegRNN\cite{lin2023segrnn}) is trained as an independent supervised model on each of the 20 datasets. 
This makes our evaluation protocol substantially more demanding in terms of cross-dataset generalization.

As shown in Fig.~\ref{experiment1}(b),(d), on the seen datasets the unified model achieves the lowest or near-lowest MAE, RMSE, and MAPE on most datasets, and consistently exhibits a tighter error distribution within each dataset. 
Using the cell-wise average errors as inputs to a Friedman test (\textbf{Supplementary Algorithm} 3), we further compare the average ranks of all methods across datasets. 
The results indicate that the proposed method attains rank~1 on a substantial fraction of datasets, while its worst rank on the remaining datasets never falls below the top three. 
These findings suggest that, even under the same training information as the baselines, the unified TSFM model maintains stable and superior relative performance across diverse chemistries, temperature ranges, and capacity scales.

The results on unseen datasets, shown in Fig.~\ref{experiment1}(a),(c), more clearly highlight the zero-shot generalization capability of the model. 
EOCV2 contains high–energy-density NMC cells with SiO-containing anodes; the SJTU dataset comprises large-capacity (13~Ah) LFP/NMC cylindrical cells; WZU covers LFP cells and parallel modules subjected to seasonally varying ambient temperatures; and Cambridge consists of LCO coin cells with capacities of only several tens of mAh. 
These datasets therefore differ markedly from the training corpus in terms of chemistry, form factor, and capacity scale. 
Despite being completely unseen during training, our unified model achieves MAE, RMSE, and MAPE values on these four datasets that are comparable to, and in many cases better than, those of per-dataset supervised baselines. 
Consistently, the Friedman average-rank analysis on the unseen datasets shows that our method most frequently attains rank~1, with its worst rank not exceeding~3.

Overall, these results demonstrate that the proposed TSFM framework can effectively forecast capacity degradation across chemistries, temperature ranges, and capacity scales while training only a single model. 
Even though competing methods require retraining a separate model for each dataset, our approach maintains competitive or superior performance on both seen and unseen datasets, supporting the feasibility of large time-series foundation models as a unified backbone for real-world battery health forecasting.
The experimental plan and results of the longer horizon study can be found in \textbf{Supplementary note}~7.

\subsection*{Interpretation maps for degradation time-series forecasting.}

\begin{figure*}[!htbp]
  \centering
  \panel[0.47\linewidth]{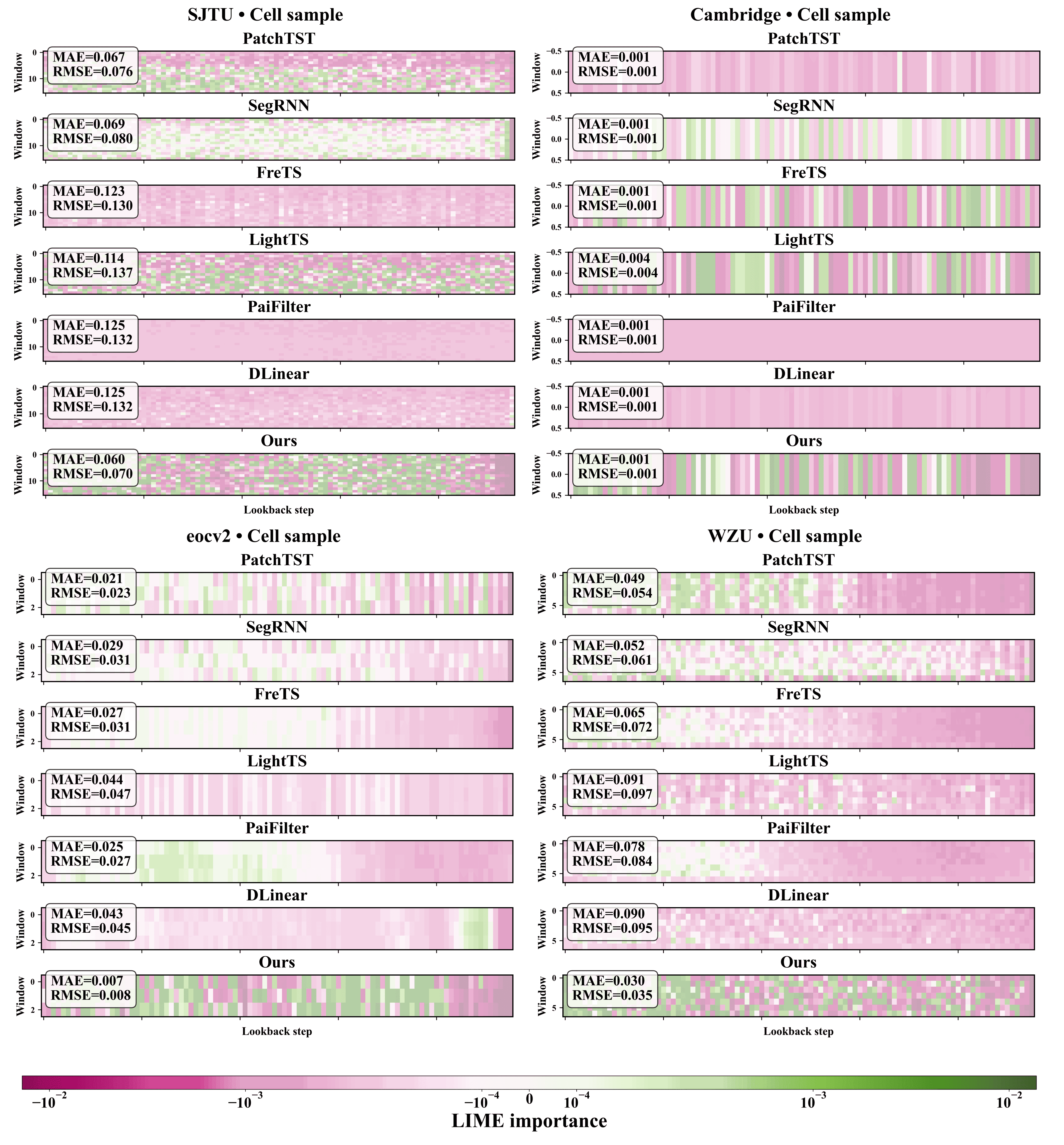}{(a)}\hfill
  \panel[0.47\linewidth]{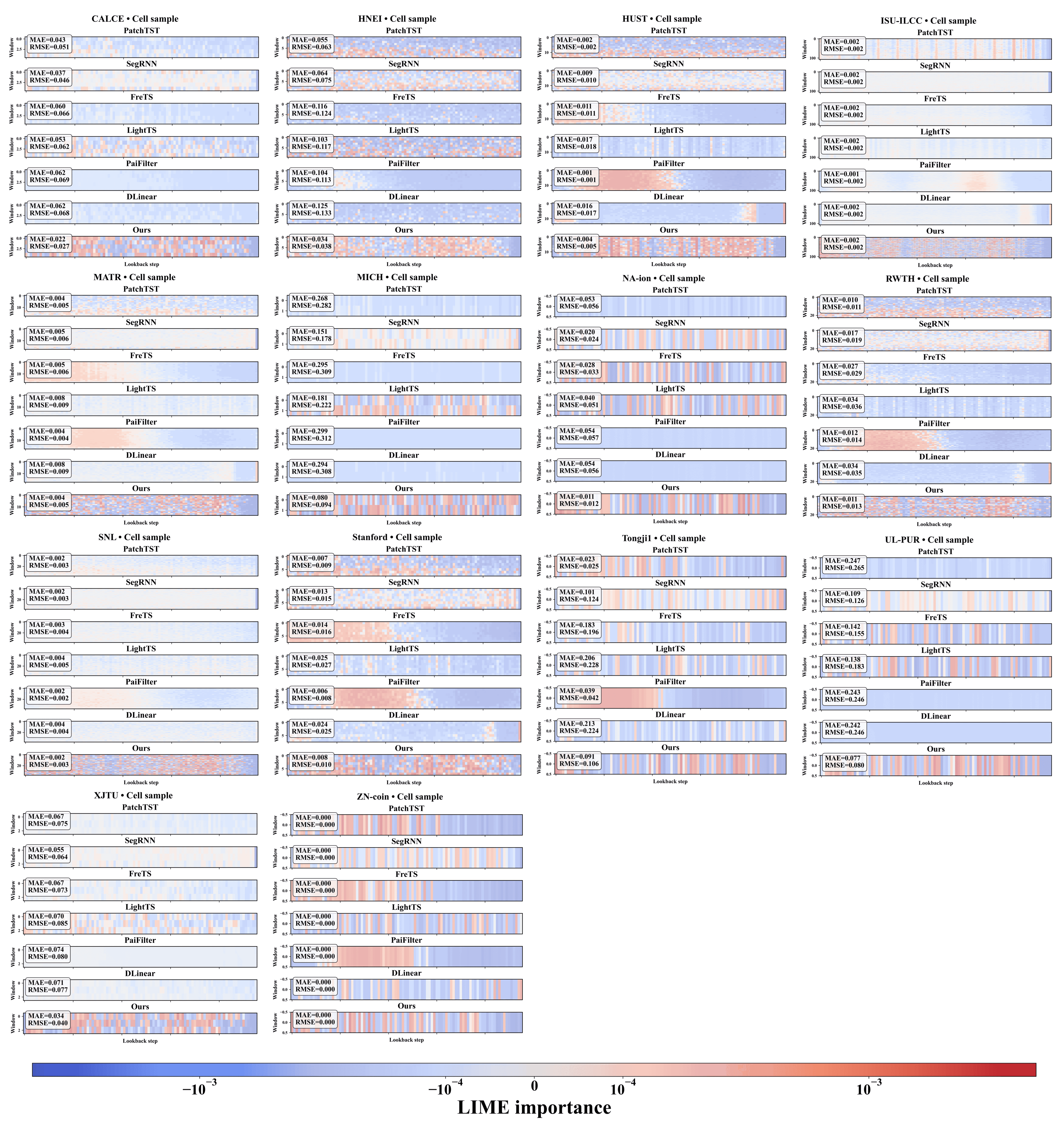}{(b)}\\
    \panel[\linewidth]{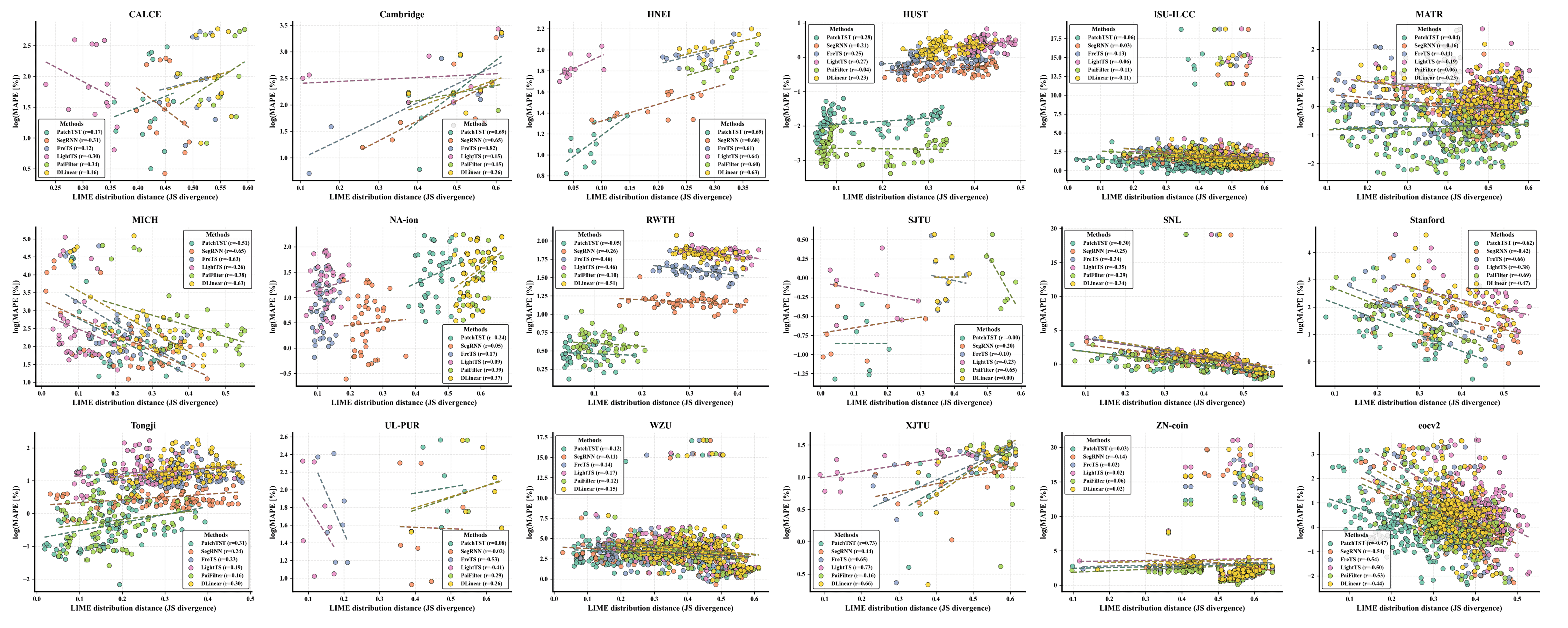}{(c)}\hfill
  \caption{\textbf{LIME-based interpretation of degradation forecasting.} 
  \textbf{(a)} Comparison of LIME importance maps for the proposed method and competing
baselines on unseen datasets. 
  \textbf{(b)} Comparison of LIME importance maps for the proposed method and competing
baselines on seen datasets.
  \textbf{(c)} Relationship between explanation discrepancy and forecasting accuracy:
scatter plot of LIME distribution distance versus $\log(\mathrm{MAPE})$ for all
methods, linking differences in attribution patterns to model performance.}
  \label{experiment2}
\end{figure*}

The degradation–feature analysis in Fig.~\ref{data}(b) shows that, despite clear differences in chemistry, temperature, and capacity scale, the 35 dimensionless degradation features (DFs) exhibit partially overlapping distributions and comparable effect sizes across LCO, NMC, and LFP cells. 
This suggests the existence of intrinsic, endogenous commonalities in capacity–fade trajectories that transcend individual chemistries. 
The fact that a \emph{single} TSFM-based model, trained once on the heterogeneous corpus, can generalize across such a broad range of datasets provides a model-level confirmation of this hypothesis: instead of memorizing dataset-specific artefacts, the model appears to exploit shared structural patterns in battery degradation.

To further probe what the model has learned, we conduct an interpretability study based on time-series LIME attribution maps generated by \textbf{Supplementary note} 8, \textbf{Algorithm} 4. 
For each sliding window, LIME perturbs the 96-step lookback sequence and fits a local surrogate regressor to attribute the forecast to individual time steps; 
aggregating these local explanations over all windows yields a two-dimensional ``LIME map'' whose horizontal axis corresponds to lookback positions and whose vertical axis indexes windows, with color encoding the signed importance (see the supplementary note on LIME map generation for details). Positive values highlight time steps that locally increase the predicted capacity, whereas negative values indicate time steps that drive the forecast downward.

Fig.~\ref{experiment2}(a) compares the LIME maps of the proposed method and the baselines on the four unseen datasets (EOCV2, SJTU, WZU, and Cambridge). 
Unlike several baselines, which tend to assign uniformly positive importance to the most recent part of the lookback window and weak or noisy importance to earlier steps, our model produces structured patterns of positive and negative importance over specific segments of the history. 
Combined with the forecasting results in Fig.~\ref{experiment1}(a), this indicates that the unified TSFM attains its superior accuracy by selectively focusing on degradation-relevant portions of the trajectory, rather than relying solely on short-term trends. 
Fig.~\ref{experiment2}(b) presents analogous LIME maps for the 16 seen datasets. Although the attribution patterns differ across chemistries, form factors, and operating conditions, the model consistently achieves high accuracy, suggesting that it has learned distinct yet meaningful attribution modes tailored to different battery families while still leveraging the endogenous commonalities revealed by the DF analysis.

To quantitatively assess how these attribution patterns relate to predictive performance, we analyze the relationship between explanation discrepancy and forecast error in \textbf{Supplementary note} 9. 
For each dataset–method–cell combination, we compute a \emph{LIME distribution distance} by flattening the corresponding LIME maps and measuring the Jensen–Shannon (JS) divergence between the baseline's importance distribution and that of our method; the associated prediction error is summarized by $\log(\mathrm{MAPE})$. 
The exact definitions of the JS divergence and $\log(\mathrm{MAPE})$ used here are provided in the supplementary note on LIME-based quantitative analysis. As illustrated in Fig.~\ref{experiment2}(c), the scatter plots reveal that on several representative datasets such as Cambridge, HNEI, and XJTU, larger JS divergence from our attribution map tends to correlate with higher $\log(\mathrm{MAPE})$: baselines whose time-series LIME maps more closely resemble those of the proposed method typically exhibit lower prediction errors. 
Fig.~\ref{experiment2}(d) summarizes these trends across all datasets. 
Taken together with the DF-based evidence in Fig.~\ref{data}(b), these results indicate that our unified TSFM not only delivers accurate cross-chemistry forecasts, but also learns attribution patterns that faithfully capture key aspects of the underlying capacity–degradation process, providing a plausible and physically meaningful reference for interpreting and evaluating other forecasting models.

%% file: manuscript/Tables/datasets.tex
\begin{table*}[htbp]
\centering
\scriptsize
\caption{Battery Datasets Grouped by Material System}
\label{tab:dataset_summary_reclassified}

\begingroup
\sisetup{
  detect-all,
  group-separator = {,},
  group-minimum-digits = 4,
  input-ignore = {,},
  table-number-alignment = center
}
\setlength{\tabcolsep}{3pt}
\renewcommand{\arraystretch}{1.05}

\begin{tabular}{
  @{}l  l  l
  S[table-format=3.0]
  S[table-format=7.0]
  l  l  l@{}
}
\Xhline{1.2pt}
\textbf{Dataset} & \textbf{Cathode} & \textbf{Anode} &
\textbf{Cell Num} & \textbf{Cycle Num} & \textbf{Capacity (Ah)} &
\textbf{Temperature (\si{\celsius})} & \textbf{Format} \\
\hline
\multicolumn{8}{@{}l@{}}{%
  \begingroup\setlength{\fboxsep}{0pt}\colorbox{gray!10}{\makebox[0.82\linewidth][l]{\textbf{NMC family}}}\endgroup
}\\[-0.35ex]
RWTH \cite{li2021one}         & NMC       & Carbon               & 46  & 100,423   & 3.00 & 25               & 18650 cyl. \\
ISU\_ILCC \cite{li2024predicting}   & NMC       & Graphite             & 240 & 1,194,846 & 0.25 & 30               & 502030 pouch \\
MICH\_EXP \cite{weng2021predicting}     & NMC111    & Graphite             & 18  & 6,545     & 5.00 & -5 / 25 / 45     & Pouch \\
MICH \cite{weng2021predicting}           & NMC111    & Graphite             & 40  & 19,895    & 2.36 & 25 / 45          & Pouch \\
Stanford\_2 \cite{cui2024data}   & NMC532    & Artificial Graphite  & 181 & 188,621   & 0.25 & 30               & 18650 cyl. \\
Stanford  \cite{cui2024data}    & NMC532    & Artificial Graphite  & 41  & 39,329    & 0.24 & 30               & 18650 cyl. \\
XJTU  \cite{cui2024data}         & NMC532    & Graphite             & 23  & 6,938     & 2.00 & 20               & 18650 cyl. \\
EOCV2 \cite{luh2024comprehensive}        & NMC       & Carbon--SiO          & 228 & 1,286,612 & 3.00 & 0 / 10 / 25 / 40 & 18650 cyl. \\

\multicolumn{8}{@{}l@{}}{%
  \begingroup\setlength{\fboxsep}{0pt}\colorbox{gray!10}{\makebox[0.82\linewidth][l]{\textbf{LCO family}}}\endgroup
}\\[-0.35ex]
CALCE \cite{xing2013ensemble}        & LCO       & Graphite             & 13  & 14,299    & 1.10 & 25               & Prismatic \\
Cambridge \cite{zhang2020identifying}   & LCO       & Graphite             & 10  & 2,670     & 0.045& 25 / 35 / 45     & LR2032 coin \\

\multicolumn{8}{@{}l@{}}{%
  \begingroup\setlength{\fboxsep}{0pt}\colorbox{gray!10}{\makebox[0.82\linewidth][l]{\textbf{LFP family}}}\endgroup
}\\[-0.35ex]
HUST \cite{ma2022real}         & LFP       & Graphite             & 77  & 146,120   & 1.10 & 30               & 18650 cyl. \\
MATR  \cite{severson2019data}        & LFP       & Graphite             & 168 & 139,313   & 1.10 & 30               & 18650 cyl. \\
WZU  \cite{lyu2024battery}          & LFP       & Graphite             & 242 & 106,060   & 1 / 0.8 / 3 & Seasonal  & Cells + parallel packs \\

\multicolumn{8}{@{}l@{}}{%
  \begingroup\setlength{\fboxsep}{0pt}\colorbox{gray!10}{\makebox[0.82\linewidth][l]{\textbf{Na-ion family}}}\endgroup
}\\[-0.35ex]
NA-ion \cite{tan2025batterylife}        & Unknown   & Unknown              & 64  & 12,604    & 1.00 & 25 / 30          & 18650 cyl. \\

\multicolumn{8}{@{}l@{}}{%
  \begingroup\setlength{\fboxsep}{0pt}\colorbox{gray!10}{\makebox[0.82\linewidth][l]{\textbf{Zn-ion family}}}\endgroup
}\\[-0.35ex]
ZN-coin \cite{tan2025batterylife}  & Zn & MnO\textsubscript{2}      & 91  & 40,417    & \num{5.5e-4} & 25 / 30 / 40 & Coin \\

\multicolumn{8}{@{}l@{}}{%
  \begingroup\setlength{\fboxsep}{0pt}\colorbox{gray!10}{\makebox[0.82\linewidth][l]{\textbf{hybrid chemistries}}}\endgroup
}\\[-0.35ex]
UL\_PUR \cite{juarez2020degradation}       & LCO + NCM442 & Graphite            & 10  & 2,245    & 3.40 & 23               & 18650 cyl. \\
HNEI \cite{devie2018intrinsic}        & LCO + NCM442 & Graphite            & 14  & 15,164   & 2.80 & 25               & 18650 cyl. \\
Tongji \cite{zhu2022data}       & Mixed NCA/NCM & Graphite / Graphite+Si & 130 & 59,038 & 3.5 / 2.5 & Various     & 18650 cyl. \\

\multicolumn{8}{@{}l@{}}{%
  \begingroup\setlength{\fboxsep}{0pt}\colorbox{gray!10}{\makebox[0.82\linewidth][l]{\textbf{Multi chemistry system}}}\endgroup
}\\[-0.35ex]
SNL \cite{preger2020degradation}           & LFP/NCA/NCM  & Graphite / Graphite+Si & 60  & 92,856  & 1.1 / 3.2 / 3.0 & 15 / 25 / 35 & 18650 cyl. \\
SJTU \cite{wu2025battery}          & LFP/NMC      & Graphite            & 8   & 487,200  & 13.0 & Room Temp        & IFR32135 \\
\Xhline{1.2pt}
\end{tabular}

\endgroup
\end{table*}

%% file: manuscript/discussion.tex
\section*{Discussion}

Batteries, as the core units of electrochemical energy storage systems, have become critical infrastructure for electrified transportation, renewable power integration, and distributed energy systems. 
With the rapid growth of electric vehicles and grid-scale storage, battery systems are evolving at unprecedented speed toward increasingly diverse chemistries, form factors, and operating conditions. This diversification imposes stringent requirements on battery management systems, which must provide reliable health prognostics across heterogeneous fleets rather than for a single, well-controlled laboratory cell type. 
In this work, we address this challenge by constructing a large, unified corpus of capacity degradation trajectories and developing a single TSFM-based forecasting framework that is trained once on the heterogeneous corpus and then evaluated across both seen and unseen chemistries, capacities, temperatures, and cycling protocols.

A key contribution of this study lies in the design of the degradation corpus and the associated evaluation protocol. 
We curate and harmonize twenty public aging datasets into a large-scale, multi-chemistry time-series corpus, spanning lithium-ion (LCO, NMC/NCA, LFP), sodium-ion, zinc-ion, and mixed-chemistry systems, as well as cylindrical, pouch, prismatic, and coin-cell formats over a wide temperature range and diverse test protocols. 
On top of this corpus, we adopt a window-based validation strategy and systematically assess the proposed framework along two axes: \emph{seen} versus \emph{unseen} data. 
In the seen setting, the model is trained and tested on trajectories drawn from the same datasets and chemistry families, providing a measure of within-domain performance. 
In the unseen setting, entire datasets are held out from training---for example, sodium-ion cells with previously unseen chemistry, large-format IFR32135 cells with capacities orders of magnitude higher than typical laboratory cells, and mixed-chemistry battery systems—and are only introduced at validation time. 
This design yields a strict test of zero-shot and cross-domain generalization: the model must extrapolate its learned temporal priors to chemistries, capacity scales, and operating regimes that are not represented in its training distribution.

From an application perspective, the results demonstrate that the proposed framework offers several advantages for real-world deployment. 
Across a broad collection of datasets, the unified TSFM with LoRA-based parameter-efficient adaptation matches or outperforms strong task-specific baselines that are individually trained on each dataset, while maintaining stable accuracy on chemistries and operating conditions that are completely absent during training. 
These findings suggest that, when supported by a sufficiently diverse degradation corpus and equipped with lightweight fine-tuning mechanisms, TSFM-based models can serve as reusable forecasting backbones for cross-platform battery health management. 
Looking ahead, further enriching the capacity degradation corpus with field data, emerging chemistries, and multi-scale aging experiments, together with broader validation and integration in practical BMS workflows, will be crucial to fully unlocking the application potential of TSFM-based foundation models for future battery technologies.

%% file: manuscript/method.tex
\section*{Methods}

\subsection*{Problem formulation}

In practical battery health management, operators are primarily interested in forecasting how the \textbf{discharge capacity} of each cell will degrade over future cycles under unknown or changing operating conditions. 
We therefore treat the cycle-wise discharge capacity as the \textbf{endogenous variable}, and use physically meaningful descriptors extracted from each charge--discharge cycle as \textbf{exogenous variables} that are available during training but may be missing or partially observed at deployment.

Formally, let the endogenous series be given in Eq.~\eqref{method_eq_y}.
\begin{equation}\label{method_eq_y}
y_{1:T} = \{ y_1, y_2, \ldots, y_T \} \in \mathbb{R}^{T}.
\end{equation}
Here, $y_t$ denotes the normalized discharge capacity at the $t$-th cycle.
For each cycle, we also compute $C$ physical descriptors (e.g., charging time, the mean and variance of current and voltage, voltage kurtosis), organized into the multivariate exogenous sequence in Eq.~\eqref{method_eq_X}.
\begin{equation}\label{method_eq_X}
\mathbf{X}_{1:T_x} = \left\{ \mathbf{x}_1, \mathbf{x}_2, \ldots, \mathbf{x}_{T_x} \right\}
\in \mathbb{R}^{T_x \times C}.
\end{equation}
Here, $\mathbf{x}_t = [x_t^{(1)}, \ldots, x_t^{(C)}]^\top$ collects the physical features at cycle $t$. 
In general, the look-back windows for endogenous and exogenous variables may differ, i.e., $T_x \neq T$, which covers asynchronous sampling and partially observed scenarios.

Given a historical window of $T$ cycles and its associated exogenous information, our goal is to learn a forecasting function $f_\theta$, parameterized by $\theta$, that predicts the future capacity over a horizon of length $S$, as formalized in Eq.~\eqref{method_eq1}.
\begin{equation}\label{method_eq1}
\hat{y}_{T+1:T+S} = f_\theta(y_{1:T}, \mathbf{X}_{1:T_x}).
\end{equation}
In this work, $f_\theta$ is instantiated as a time-series foundation model built on the \textbf{Timer} architecture, and is trained with both a supervised forecasting objective and a physics-informed contrastive objective so as to generalize across chemistries, form factors, and operating conditions.

\subsection*{Timer-based time-series foundation model}

To support capacity forecasting across heterogeneous and potentially unseen operating profiles, we adopt \textbf{Timer}~\cite{liu2024timer} as the backbone time-series foundation model. 
Timer is a scalable, decoder-only Transformer originally developed for multivariate time series and designed to capture long-range temporal dependencies through an autoregressive generation paradigm similar to that used in LLMs. 
In our setting, Timer provides a shared \emph{foundation} that encodes degradation dynamics and can be adapted to different cells via lightweight fine-tuning (e.g., LoRA).

Given an input sequence, we follow the original Timer formulation and partition the series into non-overlapping temporal tokens of length $S$, as shown in Eq.~\eqref{method_eq2}.
\begin{equation}\label{method_eq2}
s_i = \{y_{(i-1)S+1}, \ldots, y_{iS}\} \in \mathbb{R}^{S \times 1}, \quad i = 1, \ldots, N.
\end{equation}
Here, $NS$ is the total sequence length and $T = NS$. 
Each token $s_i$ is projected into a latent embedding space and augmented with temporal encoding, then passed through a stack of causal Transformer blocks, as described in Eq.~\eqref{method_eq3}.
\begin{equation}\label{method_eq3}
\begin{aligned}
&h_i^0 = W_e s_i + \text{TE}_i, \\
&h^l = \text{TrmBlock}(h^{l-1}), \quad l = 1, \ldots, L, \\
&\hat{s}_{i+1} = W_d h_i^L,
\end{aligned}
\end{equation}
where $W_e, W_d \in \mathbb{R}^{D \times (S \cdot C)}$ are the input/output projection matrices, $h_i^L$ is the final latent representation at step $i$ after $L$ decoder blocks, and $\text{TE}_i$ denotes the temporal encoding.

The model is trained with a generative forecasting objective that minimizes the mean squared error (MSE) between the predicted and true segments, as defined in Eq.~\eqref{method_eq4}.
\begin{equation}\label{method_eq4}
\mathcal{L}_{\text{forecasting}} = \frac{1}{NS} \sum_{i=2}^N \|s_i - \hat{s}_i\|_2^2.
\end{equation}
In our application, this objective teaches the foundation model to reproduce the capacity degradation trajectory in an autoregressive manner, using only the capacity history as input. 
Crucially, the same Timer backbone is also exposed to physical features through a dedicated physics-embedding pathway during training, which we describe next.

\subsection*{Physics-embedded contrastive learning}

Purely supervised training on capacity trajectories can lead to models that fit dataset-specific patterns but fail to generalize across chemistries, protocols, and environments. 
At the same time, fully conditioning on all physical variables at inference time is often impractical in real-world BMS deployments, where some signals may be noisy, missing, or unavailable. 
To bridge this gap, we introduce a \emph{physics-embedded} contrastive learning scheme: physical descriptors are used as an auxiliary training signal to shape the latent representation of the foundation model, but are not required as inputs at inference.

\paragraph{Triplet construction from physical descriptors.}
From the exogenous sequence $\mathbf{X}_{1:T_x}$, we construct fixed-length segments
of length $L$ and form training triplets
$(\mathbf{x}_{\text{anchor}}, \mathbf{x}_{\text{positive}},
\mathbf{x}_{\text{negative}})$, where
\begin{itemize}
    \item \textbf{Anchor} $\mathbf{x}_{\text{anchor}} \in \mathbb{R}^{L \times C}$
    is a segment of physical features corresponding to a given capacity window;
    \item \textbf{Positive} $\mathbf{x}_{\text{positive}} \in \mathbb{R}^{L \times C}$
    is drawn from a nearby time window (e.g., a future window offset by $L$ cycles),
    which typically reflects a similar degradation stage under slightly different
    operating conditions;
    \item \textbf{Negative} $\mathbf{x}_{\text{negative}} \in \mathbb{R}^{L \times C}$
    is sampled from a distant part of the trajectory (e.g., early vs.\ late life),
    representing a substantially different degradation state.
\end{itemize}
This triplet construction encodes the inductive bias that degradation evolves gradually, so physically similar segments close in time should be mapped to nearby representations, while far-apart segments should be separated in the latent space.

\paragraph{Physics encoder and Timer embedding.}
Each segment $\mathbf{x} \in \mathbb{R}^{L \times C}$ is first passed through a feature expansion module (implemented as a linear MLP) that maps the $C$-dimensional physical descriptors into a $D$-dimensional embedding compatible with the Timer backbone:
\[
\mathbf{z} = \Phi(\mathbf{x}) \in \mathbb{R}^{L \times D},
\]
where $\Phi(\cdot)$ denotes the physics encoder. The resulting embeddings are then
fed into the Timer model as input embeddings (instead of capacity tokens), and the
output hidden representations are aggregated into a single vector per segment. In
this way, the same foundation model learns to encode both \emph{capacity-view} and
\emph{physics-view} sequences within a shared latent space.

\paragraph{Contrastive objective.}
To align these physics-view representations, we employ an InfoNCE-style contrastive
loss that brings the anchor and positive segments closer while pushing the negative
segment away. Let $\mathbf{z}_a$, $\mathbf{z}_p$, and $\mathbf{z}_n$ denote the
aggregated embeddings of the anchor, positive, and negative segments, respectively.
We first define the cosine similarities in Eqs.~\eqref{eq:sim_ap_sim_an}, and then specify the physics-informed contrastive loss in Eq.~\eqref{method_eq5}.
\begin{equation}\label{eq:sim_ap_sim_an}
s_{ap} = \operatorname{sim}(\mathbf{z}_a, \mathbf{z}_p),s_{an} = \operatorname{sim}(\mathbf{z}_a, \mathbf{z}_n)
\end{equation}
\begin{equation}\label{method_eq5}
\mathcal{L}_{\text{physics}} = -\log \frac{\exp(s_{ap}/\tau)}{\exp(s_{ap}/\tau) + \exp(s_{an}/\tau)}.
\end{equation}
Here, $\operatorname{sim}(\cdot,\cdot)$ denotes cosine similarity and $\tau$ is a
temperature hyperparameter. In practice, we further apply a small Gaussian
perturbation to the embeddings before computing the loss, which encourages smoother,
more robust representations. This physics-informed contrastive learning forces the
Timer backbone to organize its latent space according to degradation stage and
physical context, even though capacity itself is not explicitly used in this branch.

\subsection*{Joint training and inference}

The final training objective combines the supervised capacity-forecasting loss and
the physics-embedded contrastive loss, as given in Eq.~\eqref{method_eq6}.
\begin{equation}\label{method_eq6}
\mathcal{L}_{\text{total}} = \alpha \cdot \mathcal{L}_{\text{physics}} + \beta \cdot \mathcal{L}_{\text{forecasting}}.
\end{equation}
Here, $\alpha$ and $\beta$ are weighting coefficients. The forecasting term
$\mathcal{L}_{\text{forecasting}}$ is defined in Eq.~\eqref{method_eq4}, while
$\mathcal{L}_{\text{physics}}$ is given in Eq.~\eqref{method_eq5}.

Intuitively, the forecasting loss teaches the model to reproduce capacity
trajectories from historical capacity alone, which is exactly the information
available in many real BMS logs. The physics contrastive loss, on the other hand,
injects information from charge--discharge descriptors during training, shaping the
foundation model to respect the underlying physical similarities and differences
across chemistries, temperatures, and protocols. Because the exogenous variables are
only required in the contrastive branch, they are not needed at inference time:
deployment simply feeds past capacity measurements into the Timer backbone to obtain
future capacity forecasts. This training strategy yields a physics-aware yet
input-agnostic model that can generalize across multiple battery types, operating
conditions, and datasets using a single time-series foundation model.
For more details and parameters of the methods, please refer to \textbf{Supplementary note} 10.

%% file: manuscript/addition.tex
\section*{Data Availability} 
\vspace{-4pt}
The data that support the findings of this study are available from the corresponding authors upon reasonable request (In processing).
The code of the thesis will be made open-source after acceptance.

\section*{Acknowledgments}
This work is sponsored by the National Natural Science Foundation of China under Grant 72571178, 72471143 and 72071127.

\section*{Author contributions}

Joey Chan designed and implemented the numerical experiment protocol and led the initial manuscript drafting. 
Huan Wang conceived the overall study framework and co-led the writing of the paper. 
Haoyu Pan contributed to manuscript drafting and performed detailed checks and revisions. 
Wei Wu and Zirong Wang were responsible for the procurement of experimental equipment and the acquisition of the SJTU dataset. 
Zhen Chen provided guidance on high-level architecture design and language polishing. 
The remaining co-authors offered critical feedback and domain-specific insights throughout the study. 
The project was initiated, coordinated, and supervised by Prof.\ Xi.

\section*{Competing interests}
\noindent The authors declare no competing interests.

\section*{Materials and correspondence}
\noindent Correspondence and requests for materials should be addressed to J.C.

%% file: main.bbl
\begin{thebibliography}{35}
\expandafter\ifx\csname natexlab\endcsname\relax\def\natexlab#1{#1}\fi
\expandafter\ifx\csname bibnamefont\endcsname\relax
  \def\bibnamefont#1{#1}\fi
\expandafter\ifx\csname bibfnamefont\endcsname\relax
  \def\bibfnamefont#1{#1}\fi
\expandafter\ifx\csname citenamefont\endcsname\relax
  \def\citenamefont#1{#1}\fi
\expandafter\ifx\csname url\endcsname\relax
  \def\url#1{\texttt{#1}}\fi
\expandafter\ifx\csname urlprefix\endcsname\relax\def\urlprefix{URL }\fi
\providecommand{\bibinfo}[2]{#2}
\providecommand{\eprint}[2][]{\url{#2}}

\bibitem[{\citenamefont{Radha et~al.}(2025)}]{radha2025sustainable}
\bibinfo{author}{\bibfnamefont{R.}~\bibnamefont{Radha}} \bibnamefont{et~al.},
  \bibinfo{journal}{Scientific Reports} \textbf{\bibinfo{volume}{15}},
  \bibinfo{pages}{42771} (\bibinfo{year}{2025}).

\bibitem[{\citenamefont{Link et~al.}(2025)\citenamefont{Link, Schneider,
  Stephan, Weymann, and Pl{\"o}tz}}]{link2025feasibility}
\bibinfo{author}{\bibfnamefont{S.}~\bibnamefont{Link}},
  \bibinfo{author}{\bibfnamefont{L.}~\bibnamefont{Schneider}},
  \bibinfo{author}{\bibfnamefont{A.}~\bibnamefont{Stephan}},
  \bibinfo{author}{\bibfnamefont{L.}~\bibnamefont{Weymann}}, \bibnamefont{and}
  \bibinfo{author}{\bibfnamefont{P.}~\bibnamefont{Pl{\"o}tz}},
  \bibinfo{journal}{Nature Energy} pp. \bibinfo{pages}{1--9}
  (\bibinfo{year}{2025}).

\bibitem[{\citenamefont{{Grand View Research}}(2024)}]{grandview2024market}
\bibinfo{author}{\bibnamefont{{Grand View Research}}},
  \emph{\bibinfo{title}{Battery market size, share \& trends analysis report by
  product, by application, by region, and segment forecasts, 2024 - 2030}},
  \bibinfo{howpublished}{\url{https://www.grandviewresearch.com/industry-analysis/battery-market}}
  (\bibinfo{year}{2024}), \bibinfo{note}{accessed: 2025-06-18}.

\bibitem[{\citenamefont{Cong et~al.}(2024)\citenamefont{Cong, Wang, and
  Wang}}]{cong2024review}
\bibinfo{author}{\bibfnamefont{L.}~\bibnamefont{Cong}},
  \bibinfo{author}{\bibfnamefont{W.}~\bibnamefont{Wang}}, \bibnamefont{and}
  \bibinfo{author}{\bibfnamefont{Y.}~\bibnamefont{Wang}},
  \bibinfo{journal}{Journal of Energy Storage} \textbf{\bibinfo{volume}{94}},
  \bibinfo{pages}{112406} (\bibinfo{year}{2024}).

\bibitem[{\citenamefont{Dini et~al.}(2024)\citenamefont{Dini, Colicelli, and
  Saponara}}]{dini2024review}
\bibinfo{author}{\bibfnamefont{P.}~\bibnamefont{Dini}},
  \bibinfo{author}{\bibfnamefont{A.}~\bibnamefont{Colicelli}},
  \bibnamefont{and} \bibinfo{author}{\bibfnamefont{S.}~\bibnamefont{Saponara}},
  \bibinfo{journal}{Batteries} \textbf{\bibinfo{volume}{10}},
  \bibinfo{pages}{34} (\bibinfo{year}{2024}).

\bibitem[{\citenamefont{Zou et~al.}(2024)\citenamefont{Zou, Chernyaev, Ossama,
  Seisko, and Lundstr{\"o}m}}]{zou2024leaching}
\bibinfo{author}{\bibfnamefont{Y.}~\bibnamefont{Zou}},
  \bibinfo{author}{\bibfnamefont{A.}~\bibnamefont{Chernyaev}},
  \bibinfo{author}{\bibfnamefont{M.}~\bibnamefont{Ossama}},
  \bibinfo{author}{\bibfnamefont{S.}~\bibnamefont{Seisko}}, \bibnamefont{and}
  \bibinfo{author}{\bibfnamefont{M.}~\bibnamefont{Lundstr{\"o}m}},
  \bibinfo{journal}{Scientific Reports} \textbf{\bibinfo{volume}{14}},
  \bibinfo{pages}{10818} (\bibinfo{year}{2024}).

\bibitem[{\citenamefont{Liu et~al.}(2019)\citenamefont{Liu, Zhu, and
  Cui}}]{liu2019challenges}
\bibinfo{author}{\bibfnamefont{Y.}~\bibnamefont{Liu}},
  \bibinfo{author}{\bibfnamefont{Y.}~\bibnamefont{Zhu}}, \bibnamefont{and}
  \bibinfo{author}{\bibfnamefont{Y.}~\bibnamefont{Cui}},
  \bibinfo{journal}{Nature Energy} \textbf{\bibinfo{volume}{4}},
  \bibinfo{pages}{540} (\bibinfo{year}{2019}).

\bibitem[{\citenamefont{Lu et~al.}(2023)\citenamefont{Lu, Xiong, Tian, Wang,
  and Sun}}]{lu2023deep}
\bibinfo{author}{\bibfnamefont{J.}~\bibnamefont{Lu}},
  \bibinfo{author}{\bibfnamefont{R.}~\bibnamefont{Xiong}},
  \bibinfo{author}{\bibfnamefont{J.}~\bibnamefont{Tian}},
  \bibinfo{author}{\bibfnamefont{C.}~\bibnamefont{Wang}}, \bibnamefont{and}
  \bibinfo{author}{\bibfnamefont{F.}~\bibnamefont{Sun}},
  \bibinfo{journal}{Nature Communications} \textbf{\bibinfo{volume}{14}},
  \bibinfo{pages}{2760} (\bibinfo{year}{2023}).

\bibitem[{\citenamefont{Zhang et~al.}(2023)\citenamefont{Zhang, Wang, Jiang,
  He, Huang, Wang, Zhang, Han, Guo, He et~al.}}]{zhang2023realistic}
\bibinfo{author}{\bibfnamefont{J.}~\bibnamefont{Zhang}},
  \bibinfo{author}{\bibfnamefont{Y.}~\bibnamefont{Wang}},
  \bibinfo{author}{\bibfnamefont{B.}~\bibnamefont{Jiang}},
  \bibinfo{author}{\bibfnamefont{H.}~\bibnamefont{He}},
  \bibinfo{author}{\bibfnamefont{S.}~\bibnamefont{Huang}},
  \bibinfo{author}{\bibfnamefont{C.}~\bibnamefont{Wang}},
  \bibinfo{author}{\bibfnamefont{Y.}~\bibnamefont{Zhang}},
  \bibinfo{author}{\bibfnamefont{X.}~\bibnamefont{Han}},
  \bibinfo{author}{\bibfnamefont{D.}~\bibnamefont{Guo}},
  \bibinfo{author}{\bibfnamefont{G.}~\bibnamefont{He}}, \bibnamefont{et~al.},
  \bibinfo{journal}{Nature Communications} \textbf{\bibinfo{volume}{14}},
  \bibinfo{pages}{5940} (\bibinfo{year}{2023}).

\bibitem[{\citenamefont{Tan et~al.}(2025)\citenamefont{Tan, Hong, Tang, Lu, Ma,
  Zheng, Li, Huang, and Zhang}}]{tan2025batterylife}
\bibinfo{author}{\bibfnamefont{R.}~\bibnamefont{Tan}},
  \bibinfo{author}{\bibfnamefont{W.}~\bibnamefont{Hong}},
  \bibinfo{author}{\bibfnamefont{J.}~\bibnamefont{Tang}},
  \bibinfo{author}{\bibfnamefont{X.}~\bibnamefont{Lu}},
  \bibinfo{author}{\bibfnamefont{R.}~\bibnamefont{Ma}},
  \bibinfo{author}{\bibfnamefont{X.}~\bibnamefont{Zheng}},
  \bibinfo{author}{\bibfnamefont{J.}~\bibnamefont{Li}},
  \bibinfo{author}{\bibfnamefont{J.}~\bibnamefont{Huang}}, \bibnamefont{and}
  \bibinfo{author}{\bibfnamefont{T.-Y.} \bibnamefont{Zhang}}, in
  \emph{\bibinfo{booktitle}{Proceedings of the 31st ACM SIGKDD Conference on
  Knowledge Discovery and Data Mining V. 2}} (\bibinfo{year}{2025}), pp.
  \bibinfo{pages}{5789--5800}.

\bibitem[{\citenamefont{Zhu et~al.}(2022)\citenamefont{Zhu, Wang, Huang,
  Bhushan~Gopaluni, Cao, Heere, M{\"u}hlbauer, Mereacre, Dai, Liu
  et~al.}}]{zhu2022data}
\bibinfo{author}{\bibfnamefont{J.}~\bibnamefont{Zhu}},
  \bibinfo{author}{\bibfnamefont{Y.}~\bibnamefont{Wang}},
  \bibinfo{author}{\bibfnamefont{Y.}~\bibnamefont{Huang}},
  \bibinfo{author}{\bibfnamefont{R.}~\bibnamefont{Bhushan~Gopaluni}},
  \bibinfo{author}{\bibfnamefont{Y.}~\bibnamefont{Cao}},
  \bibinfo{author}{\bibfnamefont{M.}~\bibnamefont{Heere}},
  \bibinfo{author}{\bibfnamefont{M.~J.} \bibnamefont{M{\"u}hlbauer}},
  \bibinfo{author}{\bibfnamefont{L.}~\bibnamefont{Mereacre}},
  \bibinfo{author}{\bibfnamefont{H.}~\bibnamefont{Dai}},
  \bibinfo{author}{\bibfnamefont{X.}~\bibnamefont{Liu}}, \bibnamefont{et~al.},
  \bibinfo{journal}{Nature communications} \textbf{\bibinfo{volume}{13}},
  \bibinfo{pages}{2261} (\bibinfo{year}{2022}).

\bibitem[{\citenamefont{Zhang et~al.}(2020)\citenamefont{Zhang, Tang, Zhang,
  Wang, Stimming, and Lee}}]{zhang2020identifying}
\bibinfo{author}{\bibfnamefont{Y.}~\bibnamefont{Zhang}},
  \bibinfo{author}{\bibfnamefont{Q.}~\bibnamefont{Tang}},
  \bibinfo{author}{\bibfnamefont{Y.}~\bibnamefont{Zhang}},
  \bibinfo{author}{\bibfnamefont{J.}~\bibnamefont{Wang}},
  \bibinfo{author}{\bibfnamefont{U.}~\bibnamefont{Stimming}}, \bibnamefont{and}
  \bibinfo{author}{\bibfnamefont{A.~A.} \bibnamefont{Lee}},
  \bibinfo{journal}{Nature communications} \textbf{\bibinfo{volume}{11}},
  \bibinfo{pages}{1706} (\bibinfo{year}{2020}).

\bibitem[{\citenamefont{Vaswani et~al.}(2017)\citenamefont{Vaswani, Shazeer,
  Parmar, Uszkoreit, Jones, Gomez, Kaiser, and
  Polosukhin}}]{vaswani2017attention}
\bibinfo{author}{\bibfnamefont{A.}~\bibnamefont{Vaswani}},
  \bibinfo{author}{\bibfnamefont{N.}~\bibnamefont{Shazeer}},
  \bibinfo{author}{\bibfnamefont{N.}~\bibnamefont{Parmar}},
  \bibinfo{author}{\bibfnamefont{J.}~\bibnamefont{Uszkoreit}},
  \bibinfo{author}{\bibfnamefont{L.}~\bibnamefont{Jones}},
  \bibinfo{author}{\bibfnamefont{A.~N.} \bibnamefont{Gomez}},
  \bibinfo{author}{\bibfnamefont{{\L}.}~\bibnamefont{Kaiser}},
  \bibnamefont{and}
  \bibinfo{author}{\bibfnamefont{I.}~\bibnamefont{Polosukhin}},
  \bibinfo{journal}{Advances in neural information processing systems}
  \textbf{\bibinfo{volume}{30}} (\bibinfo{year}{2017}).

\bibitem[{\citenamefont{Liu et~al.}(2024)\citenamefont{Liu, Zhang, Li, Huang,
  Wang, and Long}}]{liu2024timer}
\bibinfo{author}{\bibfnamefont{Y.}~\bibnamefont{Liu}},
  \bibinfo{author}{\bibfnamefont{H.}~\bibnamefont{Zhang}},
  \bibinfo{author}{\bibfnamefont{C.}~\bibnamefont{Li}},
  \bibinfo{author}{\bibfnamefont{X.}~\bibnamefont{Huang}},
  \bibinfo{author}{\bibfnamefont{J.}~\bibnamefont{Wang}}, \bibnamefont{and}
  \bibinfo{author}{\bibfnamefont{M.}~\bibnamefont{Long}},
  \bibinfo{journal}{arXiv preprint arXiv:2402.02368}  (\bibinfo{year}{2024}).

\bibitem[{\citenamefont{Li et~al.}(2021)\citenamefont{Li, Sengupta, Dechent,
  Howey, Annaswamy, and Sauer}}]{li2021one}
\bibinfo{author}{\bibfnamefont{W.}~\bibnamefont{Li}},
  \bibinfo{author}{\bibfnamefont{N.}~\bibnamefont{Sengupta}},
  \bibinfo{author}{\bibfnamefont{P.}~\bibnamefont{Dechent}},
  \bibinfo{author}{\bibfnamefont{D.}~\bibnamefont{Howey}},
  \bibinfo{author}{\bibfnamefont{A.}~\bibnamefont{Annaswamy}},
  \bibnamefont{and} \bibinfo{author}{\bibfnamefont{D.~U.} \bibnamefont{Sauer}},
  \bibinfo{journal}{Journal of Power Sources} \textbf{\bibinfo{volume}{506}},
  \bibinfo{pages}{230024} (\bibinfo{year}{2021}).

\bibitem[{\citenamefont{Li et~al.}(2024)\citenamefont{Li, Zhou, Thelen, Howey,
  and Hu}}]{li2024predicting}
\bibinfo{author}{\bibfnamefont{T.}~\bibnamefont{Li}},
  \bibinfo{author}{\bibfnamefont{Z.}~\bibnamefont{Zhou}},
  \bibinfo{author}{\bibfnamefont{A.}~\bibnamefont{Thelen}},
  \bibinfo{author}{\bibfnamefont{D.~A.} \bibnamefont{Howey}}, \bibnamefont{and}
  \bibinfo{author}{\bibfnamefont{C.}~\bibnamefont{Hu}}, \bibinfo{journal}{Cell
  Reports Physical Science} \textbf{\bibinfo{volume}{5}}
  (\bibinfo{year}{2024}).

\bibitem[{\citenamefont{Weng et~al.}(2021)\citenamefont{Weng, Mohtat, Attia,
  Sulzer, Lee, Less, and Stefanopoulou}}]{weng2021predicting}
\bibinfo{author}{\bibfnamefont{A.}~\bibnamefont{Weng}},
  \bibinfo{author}{\bibfnamefont{P.}~\bibnamefont{Mohtat}},
  \bibinfo{author}{\bibfnamefont{P.~M.} \bibnamefont{Attia}},
  \bibinfo{author}{\bibfnamefont{V.}~\bibnamefont{Sulzer}},
  \bibinfo{author}{\bibfnamefont{S.}~\bibnamefont{Lee}},
  \bibinfo{author}{\bibfnamefont{G.}~\bibnamefont{Less}}, \bibnamefont{and}
  \bibinfo{author}{\bibfnamefont{A.}~\bibnamefont{Stefanopoulou}},
  \bibinfo{journal}{Joule} \textbf{\bibinfo{volume}{5}}, \bibinfo{pages}{2971}
  (\bibinfo{year}{2021}).

\bibitem[{\citenamefont{Cui et~al.}(2024)\citenamefont{Cui, Kang, Wang, Rose,
  Lian, Geslin, Torrisi, Bazant, Sun, and Chueh}}]{cui2024data}
\bibinfo{author}{\bibfnamefont{X.}~\bibnamefont{Cui}},
  \bibinfo{author}{\bibfnamefont{S.~D.} \bibnamefont{Kang}},
  \bibinfo{author}{\bibfnamefont{S.}~\bibnamefont{Wang}},
  \bibinfo{author}{\bibfnamefont{J.~A.} \bibnamefont{Rose}},
  \bibinfo{author}{\bibfnamefont{H.}~\bibnamefont{Lian}},
  \bibinfo{author}{\bibfnamefont{A.}~\bibnamefont{Geslin}},
  \bibinfo{author}{\bibfnamefont{S.~B.} \bibnamefont{Torrisi}},
  \bibinfo{author}{\bibfnamefont{M.~Z.} \bibnamefont{Bazant}},
  \bibinfo{author}{\bibfnamefont{S.}~\bibnamefont{Sun}}, \bibnamefont{and}
  \bibinfo{author}{\bibfnamefont{W.~C.} \bibnamefont{Chueh}},
  \bibinfo{journal}{Joule} \textbf{\bibinfo{volume}{8}}, \bibinfo{pages}{3072}
  (\bibinfo{year}{2024}).

\bibitem[{\citenamefont{Luh and Blank}(2024)}]{luh2024comprehensive}
\bibinfo{author}{\bibfnamefont{M.}~\bibnamefont{Luh}} \bibnamefont{and}
  \bibinfo{author}{\bibfnamefont{T.}~\bibnamefont{Blank}},
  \bibinfo{journal}{Scientific Data} \textbf{\bibinfo{volume}{11}},
  \bibinfo{pages}{1004} (\bibinfo{year}{2024}).

\bibitem[{\citenamefont{Xing et~al.}(2013)\citenamefont{Xing, Ma, Tsui, and
  Pecht}}]{xing2013ensemble}
\bibinfo{author}{\bibfnamefont{Y.}~\bibnamefont{Xing}},
  \bibinfo{author}{\bibfnamefont{E.~W.} \bibnamefont{Ma}},
  \bibinfo{author}{\bibfnamefont{K.-L.} \bibnamefont{Tsui}}, \bibnamefont{and}
  \bibinfo{author}{\bibfnamefont{M.}~\bibnamefont{Pecht}},
  \bibinfo{journal}{Microelectronics reliability}
  \textbf{\bibinfo{volume}{53}}, \bibinfo{pages}{811} (\bibinfo{year}{2013}).

\bibitem[{\citenamefont{Ma et~al.}(2022)\citenamefont{Ma, Xu, Jiang, Cheng,
  Yang, Shen, Yang, Huang, Ding, and Yuan}}]{ma2022real}
\bibinfo{author}{\bibfnamefont{G.}~\bibnamefont{Ma}},
  \bibinfo{author}{\bibfnamefont{S.}~\bibnamefont{Xu}},
  \bibinfo{author}{\bibfnamefont{B.}~\bibnamefont{Jiang}},
  \bibinfo{author}{\bibfnamefont{C.}~\bibnamefont{Cheng}},
  \bibinfo{author}{\bibfnamefont{X.}~\bibnamefont{Yang}},
  \bibinfo{author}{\bibfnamefont{Y.}~\bibnamefont{Shen}},
  \bibinfo{author}{\bibfnamefont{T.}~\bibnamefont{Yang}},
  \bibinfo{author}{\bibfnamefont{Y.}~\bibnamefont{Huang}},
  \bibinfo{author}{\bibfnamefont{H.}~\bibnamefont{Ding}}, \bibnamefont{and}
  \bibinfo{author}{\bibfnamefont{Y.}~\bibnamefont{Yuan}},
  \bibinfo{journal}{Energy \& Environmental Science}
  \textbf{\bibinfo{volume}{15}}, \bibinfo{pages}{4083} (\bibinfo{year}{2022}).

\bibitem[{\citenamefont{Severson et~al.}(2019)\citenamefont{Severson, Attia,
  Jin, Perkins, Jiang, Yang, Chen, Aykol, Herring, Fraggedakis
  et~al.}}]{severson2019data}
\bibinfo{author}{\bibfnamefont{K.~A.} \bibnamefont{Severson}},
  \bibinfo{author}{\bibfnamefont{P.~M.} \bibnamefont{Attia}},
  \bibinfo{author}{\bibfnamefont{N.}~\bibnamefont{Jin}},
  \bibinfo{author}{\bibfnamefont{N.}~\bibnamefont{Perkins}},
  \bibinfo{author}{\bibfnamefont{B.}~\bibnamefont{Jiang}},
  \bibinfo{author}{\bibfnamefont{Z.}~\bibnamefont{Yang}},
  \bibinfo{author}{\bibfnamefont{M.~H.} \bibnamefont{Chen}},
  \bibinfo{author}{\bibfnamefont{M.}~\bibnamefont{Aykol}},
  \bibinfo{author}{\bibfnamefont{P.~K.} \bibnamefont{Herring}},
  \bibinfo{author}{\bibfnamefont{D.}~\bibnamefont{Fraggedakis}},
  \bibnamefont{et~al.}, \bibinfo{journal}{Nature Energy}
  \textbf{\bibinfo{volume}{4}}, \bibinfo{pages}{383} (\bibinfo{year}{2019}).

\bibitem[{\citenamefont{Lyu et~al.}(2024)\citenamefont{Lyu, Zhang, Zio, and
  Xiang}}]{lyu2024battery}
\bibinfo{author}{\bibfnamefont{D.}~\bibnamefont{Lyu}},
  \bibinfo{author}{\bibfnamefont{B.}~\bibnamefont{Zhang}},
  \bibinfo{author}{\bibfnamefont{E.}~\bibnamefont{Zio}}, \bibnamefont{and}
  \bibinfo{author}{\bibfnamefont{J.}~\bibnamefont{Xiang}},
  \bibinfo{journal}{Cell Reports Physical Science} \textbf{\bibinfo{volume}{5}}
  (\bibinfo{year}{2024}).

\bibitem[{\citenamefont{Juarez-Robles et~al.}(2020)\citenamefont{Juarez-Robles,
  Jeevarajan, and Mukherjee}}]{juarez2020degradation}
\bibinfo{author}{\bibfnamefont{D.}~\bibnamefont{Juarez-Robles}},
  \bibinfo{author}{\bibfnamefont{J.~A.} \bibnamefont{Jeevarajan}},
  \bibnamefont{and} \bibinfo{author}{\bibfnamefont{P.~P.}
  \bibnamefont{Mukherjee}}, \bibinfo{journal}{Journal of The Electrochemical
  Society} \textbf{\bibinfo{volume}{167}}, \bibinfo{pages}{160510}
  (\bibinfo{year}{2020}).

\bibitem[{\citenamefont{Devie et~al.}(2018)\citenamefont{Devie, Baure, and
  Dubarry}}]{devie2018intrinsic}
\bibinfo{author}{\bibfnamefont{A.}~\bibnamefont{Devie}},
  \bibinfo{author}{\bibfnamefont{G.}~\bibnamefont{Baure}}, \bibnamefont{and}
  \bibinfo{author}{\bibfnamefont{M.}~\bibnamefont{Dubarry}},
  \bibinfo{journal}{Energies} \textbf{\bibinfo{volume}{11}},
  \bibinfo{pages}{1031} (\bibinfo{year}{2018}).

\bibitem[{\citenamefont{Preger et~al.}(2020)\citenamefont{Preger, Barkholtz,
  Fresquez, Campbell, Juba, Rom{\`a}n-Kustas, Ferreira, and
  Chalamala}}]{preger2020degradation}
\bibinfo{author}{\bibfnamefont{Y.}~\bibnamefont{Preger}},
  \bibinfo{author}{\bibfnamefont{H.~M.} \bibnamefont{Barkholtz}},
  \bibinfo{author}{\bibfnamefont{A.}~\bibnamefont{Fresquez}},
  \bibinfo{author}{\bibfnamefont{D.~L.} \bibnamefont{Campbell}},
  \bibinfo{author}{\bibfnamefont{B.~W.} \bibnamefont{Juba}},
  \bibinfo{author}{\bibfnamefont{J.}~\bibnamefont{Rom{\`a}n-Kustas}},
  \bibinfo{author}{\bibfnamefont{S.~R.} \bibnamefont{Ferreira}},
  \bibnamefont{and}
  \bibinfo{author}{\bibfnamefont{B.}~\bibnamefont{Chalamala}},
  \bibinfo{journal}{Journal of The Electrochemical Society}
  \textbf{\bibinfo{volume}{167}}, \bibinfo{pages}{120532}
  (\bibinfo{year}{2020}).

\bibitem[{\citenamefont{Wu et~al.}(2025)\citenamefont{Wu, Chen, Liu, Zhou, Xia,
  and Pan}}]{wu2025battery}
\bibinfo{author}{\bibfnamefont{W.}~\bibnamefont{Wu}},
  \bibinfo{author}{\bibfnamefont{Z.}~\bibnamefont{Chen}},
  \bibinfo{author}{\bibfnamefont{W.}~\bibnamefont{Liu}},
  \bibinfo{author}{\bibfnamefont{D.}~\bibnamefont{Zhou}},
  \bibinfo{author}{\bibfnamefont{T.}~\bibnamefont{Xia}}, \bibnamefont{and}
  \bibinfo{author}{\bibfnamefont{E.}~\bibnamefont{Pan}}, \bibinfo{journal}{Cell
  Reports Physical Science} \textbf{\bibinfo{volume}{6}}
  (\bibinfo{year}{2025}).

\bibitem[{\citenamefont{Wang et~al.}(2024{\natexlab{a}})\citenamefont{Wang, Wu,
  Dong, Qin, Zhang, Liu, Qiu, Wang, and Long}}]{wang2024timexer}
\bibinfo{author}{\bibfnamefont{Y.}~\bibnamefont{Wang}},
  \bibinfo{author}{\bibfnamefont{H.}~\bibnamefont{Wu}},
  \bibinfo{author}{\bibfnamefont{J.}~\bibnamefont{Dong}},
  \bibinfo{author}{\bibfnamefont{G.}~\bibnamefont{Qin}},
  \bibinfo{author}{\bibfnamefont{H.}~\bibnamefont{Zhang}},
  \bibinfo{author}{\bibfnamefont{Y.}~\bibnamefont{Liu}},
  \bibinfo{author}{\bibfnamefont{Y.}~\bibnamefont{Qiu}},
  \bibinfo{author}{\bibfnamefont{J.}~\bibnamefont{Wang}}, \bibnamefont{and}
  \bibinfo{author}{\bibfnamefont{M.}~\bibnamefont{Long}},
  \bibinfo{journal}{arXiv preprint arXiv:2402.19072}
  (\bibinfo{year}{2024}{\natexlab{a}}).

\bibitem[{\citenamefont{Wang et~al.}(2024{\natexlab{b}})\citenamefont{Wang,
  Zhai, Zhao, Di, and Chen}}]{wang2024physics}
\bibinfo{author}{\bibfnamefont{F.}~\bibnamefont{Wang}},
  \bibinfo{author}{\bibfnamefont{Z.}~\bibnamefont{Zhai}},
  \bibinfo{author}{\bibfnamefont{Z.}~\bibnamefont{Zhao}},
  \bibinfo{author}{\bibfnamefont{Y.}~\bibnamefont{Di}}, \bibnamefont{and}
  \bibinfo{author}{\bibfnamefont{X.}~\bibnamefont{Chen}},
  \bibinfo{journal}{Nature Communications} \textbf{\bibinfo{volume}{15}},
  \bibinfo{pages}{4332} (\bibinfo{year}{2024}{\natexlab{b}}).

\bibitem[{\citenamefont{Zeng et~al.}(2023)\citenamefont{Zeng, Chen, Zhang, and
  Xu}}]{zeng2023transformers}
\bibinfo{author}{\bibfnamefont{A.}~\bibnamefont{Zeng}},
  \bibinfo{author}{\bibfnamefont{M.}~\bibnamefont{Chen}},
  \bibinfo{author}{\bibfnamefont{L.}~\bibnamefont{Zhang}}, \bibnamefont{and}
  \bibinfo{author}{\bibfnamefont{Q.}~\bibnamefont{Xu}}, in
  \emph{\bibinfo{booktitle}{Proceedings of the AAAI conference on artificial
  intelligence}} (\bibinfo{year}{2023}), vol.~\bibinfo{volume}{37}, pp.
  \bibinfo{pages}{11121--11128}.

\bibitem[{\citenamefont{Yi et~al.}(2023)\citenamefont{Yi, Zhang, Fan, Wang,
  Wang, He, An, Lian, Cao, and Niu}}]{yi2023frequency}
\bibinfo{author}{\bibfnamefont{K.}~\bibnamefont{Yi}},
  \bibinfo{author}{\bibfnamefont{Q.}~\bibnamefont{Zhang}},
  \bibinfo{author}{\bibfnamefont{W.}~\bibnamefont{Fan}},
  \bibinfo{author}{\bibfnamefont{S.}~\bibnamefont{Wang}},
  \bibinfo{author}{\bibfnamefont{P.}~\bibnamefont{Wang}},
  \bibinfo{author}{\bibfnamefont{H.}~\bibnamefont{He}},
  \bibinfo{author}{\bibfnamefont{N.}~\bibnamefont{An}},
  \bibinfo{author}{\bibfnamefont{D.}~\bibnamefont{Lian}},
  \bibinfo{author}{\bibfnamefont{L.}~\bibnamefont{Cao}}, \bibnamefont{and}
  \bibinfo{author}{\bibfnamefont{Z.}~\bibnamefont{Niu}},
  \bibinfo{journal}{Advances in Neural Information Processing Systems}
  \textbf{\bibinfo{volume}{36}}, \bibinfo{pages}{76656} (\bibinfo{year}{2023}).

\bibitem[{\citenamefont{Zhang et~al.}()\citenamefont{Zhang, Zhang, Cao, Bian,
  Yi, Zheng, and Li}}]{zhang2207less}
\bibinfo{author}{\bibfnamefont{T.}~\bibnamefont{Zhang}},
  \bibinfo{author}{\bibfnamefont{Y.}~\bibnamefont{Zhang}},
  \bibinfo{author}{\bibfnamefont{W.}~\bibnamefont{Cao}},
  \bibinfo{author}{\bibfnamefont{J.}~\bibnamefont{Bian}},
  \bibinfo{author}{\bibfnamefont{X.}~\bibnamefont{Yi}},
  \bibinfo{author}{\bibfnamefont{S.}~\bibnamefont{Zheng}}, \bibnamefont{and}
  \bibinfo{author}{\bibfnamefont{J.}~\bibnamefont{Li}}, \bibinfo{journal}{arXiv
  preprint arXiv:2207.01186}  (????).

\bibitem[{\citenamefont{Yi et~al.}(2024)\citenamefont{Yi, Fei, Zhang, He, Hao,
  Lian, and Fan}}]{yi2024filternet}
\bibinfo{author}{\bibfnamefont{K.}~\bibnamefont{Yi}},
  \bibinfo{author}{\bibfnamefont{J.}~\bibnamefont{Fei}},
  \bibinfo{author}{\bibfnamefont{Q.}~\bibnamefont{Zhang}},
  \bibinfo{author}{\bibfnamefont{H.}~\bibnamefont{He}},
  \bibinfo{author}{\bibfnamefont{S.}~\bibnamefont{Hao}},
  \bibinfo{author}{\bibfnamefont{D.}~\bibnamefont{Lian}}, \bibnamefont{and}
  \bibinfo{author}{\bibfnamefont{W.}~\bibnamefont{Fan}},
  \bibinfo{journal}{Advances in Neural Information Processing Systems}
  \textbf{\bibinfo{volume}{37}}, \bibinfo{pages}{55115} (\bibinfo{year}{2024}).

\bibitem[{\citenamefont{Nie}(2022)}]{nie2022time}
\bibinfo{author}{\bibfnamefont{Y.}~\bibnamefont{Nie}}, \bibinfo{journal}{arXiv
  preprint arXiv:2211.14730}  (\bibinfo{year}{2022}).

\bibitem[{\citenamefont{Lin et~al.}(2023)\citenamefont{Lin, Lin, Wu, Zhao, Mo,
  and Zhang}}]{lin2023segrnn}
\bibinfo{author}{\bibfnamefont{S.}~\bibnamefont{Lin}},
  \bibinfo{author}{\bibfnamefont{W.}~\bibnamefont{Lin}},
  \bibinfo{author}{\bibfnamefont{W.}~\bibnamefont{Wu}},
  \bibinfo{author}{\bibfnamefont{F.}~\bibnamefont{Zhao}},
  \bibinfo{author}{\bibfnamefont{R.}~\bibnamefont{Mo}}, \bibnamefont{and}
  \bibinfo{author}{\bibfnamefont{H.}~\bibnamefont{Zhang}},
  \bibinfo{journal}{arXiv preprint arXiv:2308.11200}  (\bibinfo{year}{2023}).

\end{thebibliography}
